\documentclass[11pt]{article}
\usepackage{opendrivelab}
\usepackage{pifont}
\usepackage[table]{xcolor}
\usepackage{float}
\usepackage{multirow}
\usepackage{array}
\usepackage{tabularx}
\usepackage{ltablex}
\keepXColumns
\newcolumntype{L}[1]{>{\raggedright\arraybackslash}p{#1}}
\newcommand{\cmark}{\ding{51}} 
\newcommand{\xmark}{\ding{55}} 
\newlength{\orglogoheight}
\title{HandEdit: A Unified Benchmark for Egocentric Human-to-Robot Dexterous Hand Image Editing}

\orglabel{}
\orglogo{%
  \makebox[\linewidth][l]{%
    \includegraphics[height=1.2cm,keepaspectratio]{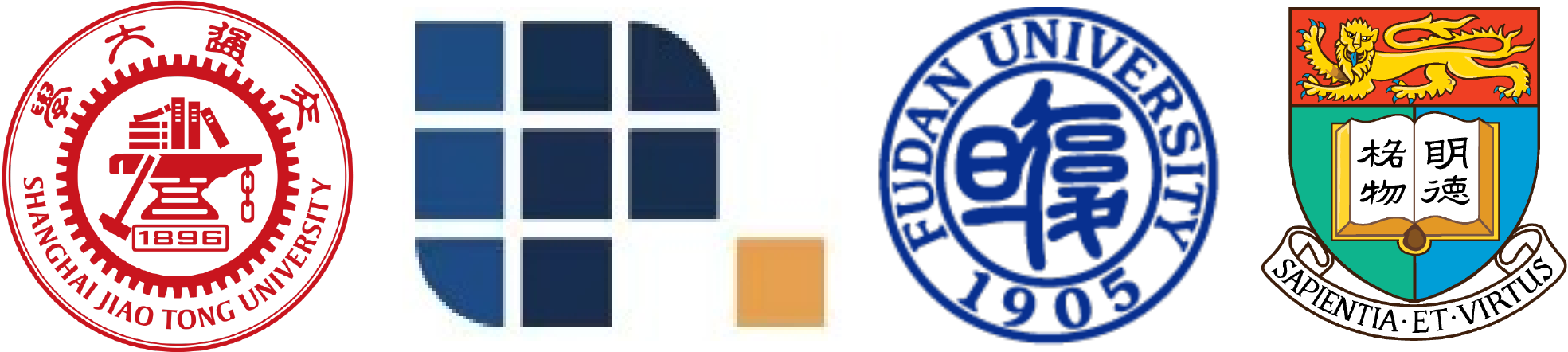}%
  }%
}

\contributors{%
  \small
  Zhenjie Yang\textsuperscript{2,3*}\quad
  Xingyu Jiao\textsuperscript{1*}\quad
  Guopeng Zhong\textsuperscript{3}\quad
  Shuzhe Yang\textsuperscript{3}\quad
  Shi Che\textsuperscript{3} \quad 
  Chao Wu\textsuperscript{1} \quad
  Chenyu Jiang\textsuperscript{1} \quad
  Dongjie Zhang\textsuperscript{1}\quad
  Yideng Zhang\textsuperscript{3}\quad
  Zheng Zhang\textsuperscript{3}\quad
  Muyun Jiang\textsuperscript{5}\quad
  Haisheng Su\textsuperscript{3}\quad
  Shuang Jin\textsuperscript{2}\quad
  Donghang Zhang\textsuperscript{2}\quad \\
  Chao Yang\textsuperscript{6}\quad
  Li Chen\textsuperscript{4}\quad 
  Hongyang Li\textsuperscript{4}\quad 
  Zuxuan Wu\textsuperscript{1}\quad 
  Yu-Gang Jiang\textsuperscript{1}\quad
  Xiaosong Jia\textsuperscript{1\dag}\quad
  Junchi Yan\textsuperscript{3\dag}\quad
}

\contriblegend{
$^{1}$ HKU \quad
}

\contriblegend{
    \textsuperscript{1} Fudan University \quad
    \textsuperscript{2} Inspire Robots \quad
    \textsuperscript{3} Shanghai Jiao Tong University \quad \\
    \textsuperscript{4} The University of Hong Kong \quad 
    \textsuperscript{5} Nanyang Technological University  \quad 
    \textsuperscript{6} Shanghai AI Laboratory \quad \\
    \textsuperscript{*}\,Equal contribution
    \textsuperscript{\dag}\,Corresponding authors \quad \\
    Contact: \href{}{\texttt{yangzhenjie@sjtu.edu.cn}, \texttt{jiaxiaosong@fudan.edu.cn}}
}

\projectlinks{%
  \link{Project Page}{https://handedit.github.io/}\quad
  \link{Full Code}{https://github.com/HandEdit/HandEdit}\quad
  \link{Dataset}{https://huggingface.co/datasets/HandEdit/HandEdit-Full}\quad
  }

\begin{document}
\maketitle
\begin{center}
    \includegraphics[width=0.95\textwidth]{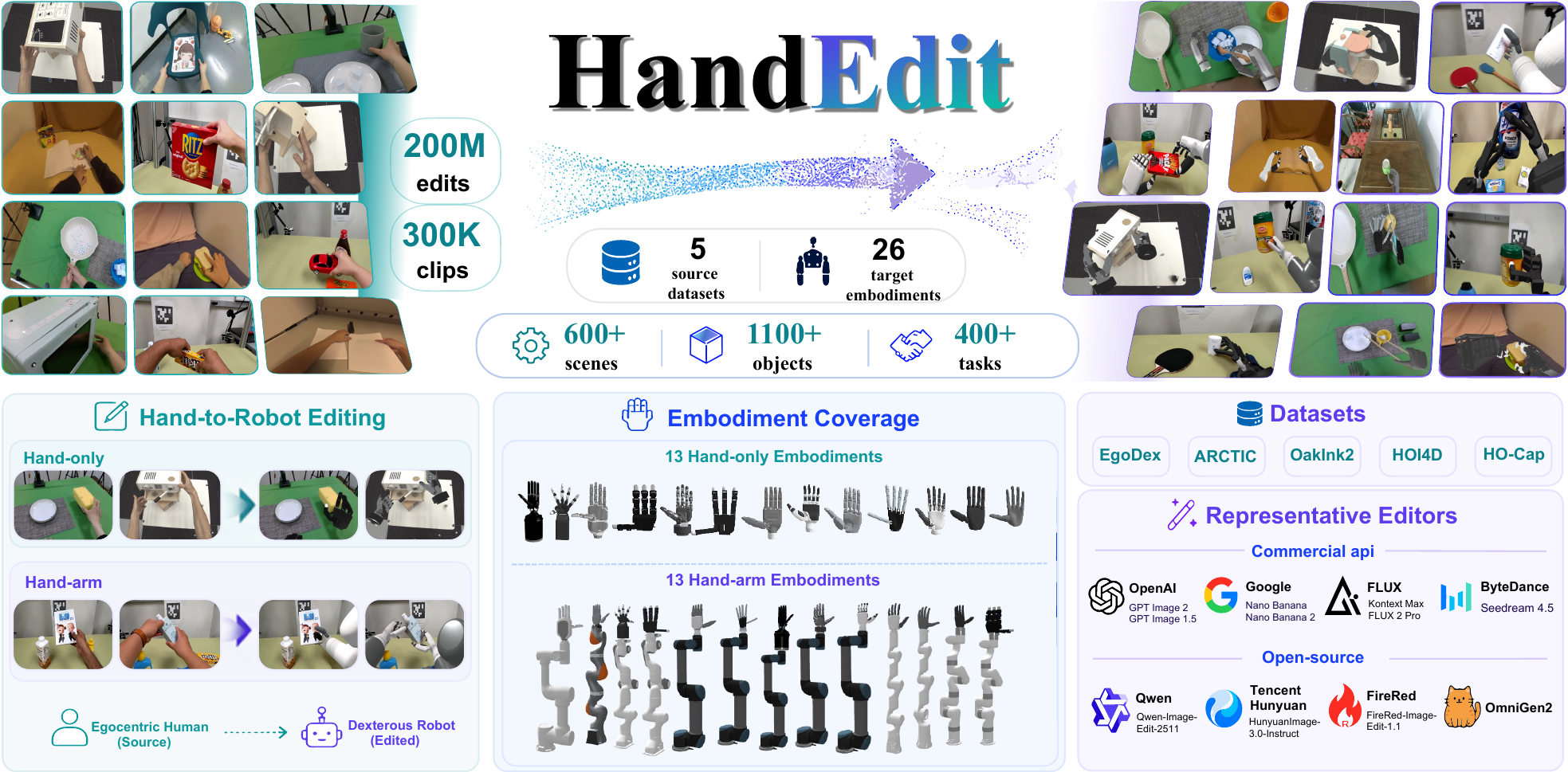}
    \captionof{figure}{\textbf{HandEdit} is an embodiment-aware image-editing dataset and benchmark for editing egocentric images of human hand and arm manipulation into diverse dexterous robotic embodiments.}
    \label{fig:teaser}
\end{center}

\begin{abstract}
\setlength{\parindent}{0pt}
Robotic manipulation with dexterous hands is a cornerstone of Embodied AI, yet its progress is stifled by the high cost of collecting embodiment-aware teleoperation data. While abundant egocentric videos of human hands offer a scalable alternative, the profound discrepancies in appearance, articulation, and  camera viewpoints between human and robotic data raise significant challenges for co-training. Though existing general image-editing models demonstrate strong capabilities, they lack necessary embodiment-specific priors to fully bridge this gap. In this work, we present~\textbf{HandEdit}, a unified large-scale embodiment-aware image-editing dataset and benchmark specifically designed to transform human hands and arms into various dexterous robotic embodiments within egocentric frames. HandEdit comprises over 200M editing instances derived from five diverse source datasets, covering 26 distinct URDFs, including 13 hand-only and 13 hand-arm configurations. Alongside the dataset, we establish a unified benchmark protocol with two tracks:~\textbf{\texttt{Hand-only}} and~\textbf{\texttt{Hand-Arm}}, supporting URDF-conditioned evaluation. We conduct extensive evaluations of 11 representative image-editing baselines using a multi-dimensional metric suite, including generic similarity metrics, VLM-based judgment, and embodiment-aware metrics. HandEdit serves as a critical resource at the intersection of image editing and robotics: it advances embodiment-aware editing models while enabling scalable dexterous robotic learning from abundant human video data, paving the way for more generalizable Embodied AI.
\end{abstract}

\section{Introduction}
Dexterous manipulation is a crucial capability in Embodied AI (EAI). Recent progress achieved in EAI has been driven by scaling up data from diverse sources, including large-scale robotic teleoperation datasets~\cite{rt1, openx, droid, bu2025agibot_arxiv, walke2023bridgedata, wu2024robomind, mobilealoha}, UMI-style demonstrations~\cite{zhaxizhuoma2025fastumi, liu2025fastumi, chi2024universal, xu2025exumi, ha2024umilegs, dexumi}, and egocentric human videos~\cite{ego4d, egodex, hoi4d, hot3d, openego, egoverse, egolife, egolive}. Among these, robotic teleoperation data offer direct robot-executable demonstrations, but is expensive for large-scale collection and typically constrained by specific hardware platforms, lab environments, and task setups. UMI-style data collection improves scalability by using portable human-operated devices, yet it still requires dedicated equipment and embodiment-specific demonstrations. In contrast, egocentric human videos are naturally abundant, diverse, and scalable, capturing rich hand-object interactions across daily activities, objects, environments, and long-horizon tasks. As a result, they provide a promising route toward scaling visual experience for dexterous manipulation.

However, directly leveraging egocentric human video for robotic policy learning remains an open challenge~\cite{egomimic, egozero, okami, egoscale}, owing to the fundamental embodiment gap between human hands and dexterous robotic hands. Human hands and robotic dexterous hands differ substantially in appearance, geometry, articulation, kinematic constraints, and interaction patterns. This gap becomes even larger for hand-arm systems, where the visible wrist and arm must also be consistent with the target robotic morphology and camera viewpoint. 

Consequently, human egocentric images cannot be converted into robot data by simple style transfer or generic appearance editing. For embodied learning, the edited image must preserve the scene, object state, task semantics, and hand-object contact. At the same time, the replaced hand or arm should form a physically plausible robotic embodiment that respects the geometry and articulation specified by the target Unified Robot Description Format (URDF), while remaining coherent with the original manipulation context.

Existing approaches~\cite{h2r, heng2025rwor, phantom, masquerade, shadow, h2rgrounder, mitty, mimicdreamer, dexumi} have explored ``robotizing'' human videos by inserting rendered robot embodiments into human scenes. While they demonstrate the value of bridging the human-robot visual embodiment gap for robotic learning, they are largely tied to parallel grippers, fixed robot platforms, or embodiment-specific pipelines. 
More fundamentally, human-to-robot dexterous hand editing has not been systematically studied as an image editing problem: \textbf{there is still no dedicated large-scale dataset, unified evaluation protocol, or comprehensive benchmark for assessing URDF-conditioned editing quality across diverse dexterous hand and hand-arm embodiments}.

To fill this gap, we introduce \textbf{HandEdit}, a dataset and benchmark for egocentric human-to-robot dexterous hand image editing. HandEdit is built from five public hand-object datasets: EgoDex~\cite{egodex}, ARCTIC~\cite{arctic}, OakInk2~\cite{oakink2}, HOI4D~\cite{hoi4d}, and HO-Cap~\cite{hocap}. HandEdit currently contains over 200M image-level editing instances derived from 300K clips, covering more than 600 scenes and 1.1K objects. It spans 26 target URDF embodiments, including 13 hand-only and 13 hand-arm configurations, as detailed in Appendix~\ref{app:embodiment_description_bank}. HandEdit further establishes a unified benchmark with two tracks, \textbf{\texttt{Hand-only}} and \textbf{\texttt{Hand-Arm}}, enabling URDF-conditioned editing quality evaluation under different settings. To support URDF-conditioned evaluation, we propose a comprehensive metric suite spanning generic similarity, VLM-based judgment, and embodiment-aware metrics, which is used to comprehensively benchmark 11 representative commercial and open-source image editors. 

\textbf{In sum, our main contributions are as follows:}
\begin{itemize}[noitemsep,topsep=0pt,leftmargin=*,itemsep=2pt]
\item We introduce \textbf{HandEdit}, the first large-scale embodiment-aware image-editing dataset tailored for egocentric human-to-robot dexterous manipulation transfer. Built upon five public egocentric manipulation datasets, HandEdit covers 26 target URDF embodiments, including 13 hand-only and 13 hand-arm configurations, and yields over 200M image-level editing instances.
\item For the first time, we establish the unified benchmark for embodiment-aware dexterous hand editing, with two tracks: \textbf{\texttt{Hand-only}} and \textbf{\texttt{Hand-Arm}}. The benchmark supports URDF-conditioned evaluation and enables systematic comparison across diverse robotic embodiments.
\item We design a comprehensive evaluation suite that combines generic visual similarity metrics, VLM-based judgment, and embodiment-aware metrics, providing a more faithful assessment of both visual quality and robotic embodiment consistency.
\item We benchmark 11 representative commercial and open-source image editors under their original conditioning interfaces, together with human evaluation for metric validation, revealing the limitations of existing general-purpose editors and highlighting HandEdit as a foundational dataset and benchmark for developing future embodiment-aware editing models.
\end{itemize}

\begin{table*}[t!]
\caption{\textbf{Comparison with Representative Editing Benchmarks and Datasets}. HandEdit is the only benchmark that supports embodiment-conditional image-editing. ``Dexterous Hand`` indicates whether the benchmark explicitly involves hand-centric or hand-object editing. ``\#Embod.'' denotes the number of supported URDF-specified robotic embodiments for editing.}\label{tab:benchmark_comparison}
\centering
\resizebox{1\textwidth}{!}{
\begin{tabular}{l|ccccccc}
\toprule
\textbf{Benchmark} & \textbf{Venue} & \textbf{Ego-centric} & \textbf{Dexterous Hand} & \textbf{    URDF-cond} & \textbf{\#Embod.} & \textbf{Scale} & \textbf{Metrics} \\
\midrule
EditBench~\cite{wang2023imagen} & CVPR'23 & \xmark & \xmark & \xmark &-- & 240 & CLIP / Human \\
InstructPix2Pix~\cite{instructpix2pix} & CVPR'23 & \xmark & \xmark & \xmark &-- & 450K & CLIP / Human \\
MagicBrush~\cite{magicbrush} & NeurIPS'23 & \xmark & \xmark & \xmark &-- & 10K & L1/L2 / CLIP / DINO / Human \\
OmniEdit~\cite{Omniedit} & ICLR'24 & \xmark & \xmark & \xmark &-- & 1.2M & VIEScore \\
Emu Edit~\cite{sheynin2024emu} & CVPR'24  & \xmark & \xmark & \xmark &-- & 3.1K & CLIP / L1 / DINO / Human \\
UltraEdit~\cite{ultraedit} & NeurIPS'24 & \xmark & \xmark & \xmark &-- & 4M & L1 / CLIP / DINO \\
SEED-Data-Edit~\cite{ge2024seed} & arXiv'24 & \xmark & \xmark & \xmark &-- & 3.7M & MLLM / Qual. \\
HQ-Edit~\cite{hui2024hq} & ICLR'25 & \xmark & \xmark & \xmark &-- & 200K & Align. / Coherence \\
AnyEdit~\cite{yu2024anyedit} & CVPR'25 & \xmark & \xmark & \xmark &-- & 2.5M & CLIP / DINO / VLM \\
VE-Bench~\cite{sun2024bench} & AAAI'25 & \xmark & \xmark & \xmark &-- & 1.17K & MOS / QA \\
MotionEdit~\cite{motionedit} & CVPR'26 & \xmark & \xmark & \xmark &-- & 10K & MLLM / MAS / Win Rate \\
EgoEdit~\cite{egoedit} & CVPR'26 & \cmark & \xmark$^*$ & \xmark &-- & 99.7K & VLM / PS / TA / TC \\
\midrule
\rowcolor{gray!15}
\textbf{HandEdit} & -- & \textbf{\cmark} & \textbf{\cmark} & \textbf{\cmark} &\textbf{26} & \textbf{200M} & \textbf{Sim. / VLM / Embodiment} \\
\bottomrule
\end{tabular}
}
\begin{flushleft}
\tiny
\textit{Note.} $^*$ EgoEdit targets general egocentric video editing with hand preservation, rather than URDF-conditioned human-to-robot hand/arm mapping for embodiments.
\end{flushleft}
\end{table*}

\section{Related Work}

\textbf{Large-Scale Data for Embodied Manipulation.}
Large-scale data has become a key driver of embodied manipulation. Robot teleoperation datasets provide direct robot-executable demonstrations across diverse tasks and embodiments~\cite{rt1, openx, droid, bu2025agibot_arxiv, walke2023bridgedata, wu2024robomind, mobilealoha}, but their collection remains costly and often tied to specific robot platforms, sensors, and environments. UMI-style systems improve scalability by enabling portable, in-the-wild demonstration collection and policy deployment across platforms~\cite{chi2024universal, zhaxizhuoma2025fastumi, liu2025fastumi, xu2025exumi, ha2024umilegs, dexumi}. However, these large-scale robot data primarily consist of gripper data; they still require dedicated interfaces or hardware adaptations, especially for dexterous hands.

\textbf{Egocentric Human-Object Interaction Datasets.}
Egocentric human videos offer a more scalable source of manipulation experience, capturing rich hand-object interactions in natural environments~\cite{ego4d, egodex, hocap, hoi4d, hot3d, openego, egoverse, egolife, egolive, arctic, oakink2, h2o}. Despite their scale and diversity, such human-centric data cannot be used directly as robot-centric visual data due to the visual, kinematic, and embodiment gaps between humans and robots.
 
\textbf{Learning Robotic Manipulation from Human Videos.}
Human videos provide scalable behavioral priors for robot manipulation~\cite{ye2026worldactionmodelszeroshot, gao2026dreamdojo, shi2026egohumanoid, egoscale}. Existing methods use such data for dexterous grasping, motion retargeting, policy learning, and robot-centric video generation. DexVIP~\cite{dexvip} learns dexterous grasping from human-object interaction videos with human hand-pose priors, while DexArt~\cite{dexart}, DexH2R~\cite{dexh2r}, and related works transfer human hand motions to dexterous robots through retargeting, simulation, or residual policy learning. Recent systems such as UniDex~\cite{unidex} and MiVLA~\cite{mivla} further scale human-video-based robot learning through foundation policies and human-robot mutual imitation. In parallel, H2R~\cite{h2r}, Phantom~\cite{phantom}, Masquerade~\cite{masquerade}, H2R-Grounder~\cite{h2rgrounder}, MimicDreamer~\cite{mimicdreamer}, Mitty~\cite{mitty}, Human2Robot~\cite{human2robot}, and DWM~\cite{dwm} convert human videos into robot-centric observations via rendering, inpainting, compositing, or video translation. While these works validate human videos as useful supervision for robot learning, they mainly target motion transfer, policy learning, video generation, or embodiment-specific pipelines. In contrast, HandEdit provides a dedicated benchmark for URDF-conditioned egocentric image editing across diverse dexterous hand and hand-arm embodiments.
 
\textbf{Image and Video Editing Methods and Benchmarks.}
Image and video editing benchmarks have progressed from general instruction-following to increasingly specialized evaluation settings. Early editing methods such as SDEdit~\cite{sdedit}, Prompt-to-Prompt~\cite{prompttoprompt}, DiffEdit~\cite{diffedit}, and ControlNet~\cite{controlnet} laid the foundation for modern instruction- and condition-guided editing. As shown in Table~\ref{tab:benchmark_comparison}, subsequent benchmarks, including InstructPix2Pix~\cite{instructpix2pix}, MagicBrush~\cite{magicbrush}, Emu Edit~\cite{sheynin2024emu}, UltraEdit~\cite{ultraedit}, OmniEdit~\cite{Omniedit}, HQ-Edit~\cite{hui2024hq}, and AnyEdit~\cite{yu2024anyedit}, provide paired data and evaluation protocols for generic image editing, typically focusing on instruction following, visual fidelity, semantic alignment, and perceptual similarity. 
More recent benchmarks move toward specialized domains: MotionEdit~\cite{motionedit} evaluates motion-centric image editing, while EgoEdit~\cite{egoedit} targets egocentric video editing under hand occlusions, hand-object interactions, and large egomotion.

As summarized in Tab.~\ref{tab:benchmark_comparison}, no prior benchmark jointly supports egocentric input, dexterous robotic hands, URDF-conditioned editing, and multi-embodiment evaluation. HandEdit fills this gap by introducing a dedicated benchmark for URDF-conditioned human-to-robot hand/arm mapping, with metrics that assess not only generic image quality but also embodiment fidelity and manipulation-context preservation.


\section{HandEdit}
HandEdit consists of five components: the data curation pipeline for generating kinematically grounded pseudo-GT targets, a multi-source dataset for egocentric human-to-robot dexterous hand image editing, a benchmark protocol with two tracks (Hand-only and Hand-Arm), comprehensive evaluation metrics, and harmonization post-processing for the final composites. We describe these components in the following section.

\begin{figure*}[t]
\centering
\includegraphics[width=\textwidth]{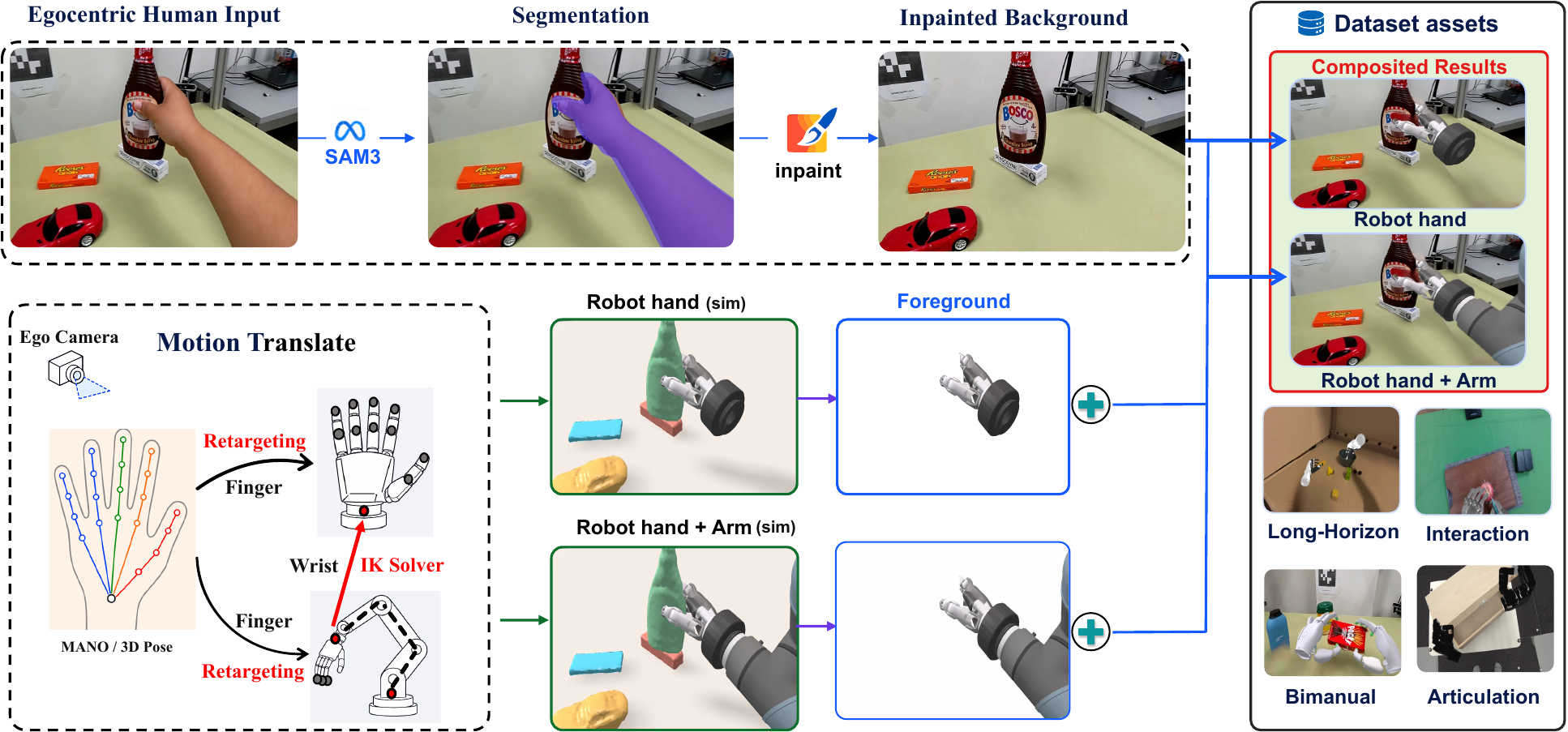}
\caption{\textbf{Data Curation Pipeline.} For the egocentric hand-object clips, HandEdit removes the human hand, recovers the occluded background, retargets the source interaction to a target URDF, and composites the rendered robot embodiment back into the scene. The composite is then harmonized to produce the final pseudo-GT target.}
\label{fig:handedit_pipeline}
\end{figure*}

\subsection{Data Curation Pipeline}
\label{sec:data_curation}
Starting from selected egocentric hand-object clips, we construct pseudo-GT targets through a staged preprocessing pipeline. First, we segment the human hand or hand-arm region using off-the-shelf segmentation models, with SAM3~\cite{sam3} serving as our default implementation, and remove the segmented foreground. We then restore the occluded background using video inpainting models, for which we use ProPainter~\cite{propainter} by default due to its strong appearance and temporal consistency on egocentric clips compared with single-image inpainting methods.

For dexterous hands, we convert the MANO~\cite{mano} or 3D hand pose into robot joint states via embodiment-specific retargeting. Following classical optimization- and geometry-based retargeting~\cite{anyteleop}, which supports both position- and vector-based modes, we first obtain a coarse alignment with a hybrid mode, accounting for embodiment differences in kinematic trees, link lengths, joint limits, and contact geometry. We then refine the target hand configuration through position-based retargeting under embodiment-specific kinematic constraints to improve local structural consistency. This preserves the local interaction layout around the manipulated object while allowing each robot hand to realize the pose according to its own kinematics.

For hand-arm embodiments, we extend the retargeted hand pose to the supporting arm. Because the physical robot base is not observable in egocentric human videos, we define a camera-relative virtual base for each source sequence and target embodiment and keep it fixed throughout the sequence. The local search is initialized from the shoulder midpoint, shoulder direction, and estimated shoulder width when shoulder keypoints are available; otherwise, it uses the wrist-trajectory center and the nominal reach of the target arm. We then perform an embodiment-adaptive local search over horizontal translation and yaw, where translational offsets are normalized by the nominal reach of the target arm. Base height, roll, and pitch are fixed according to the assumed upright mounting configuration. The human wrist trajectory is mapped to the target wrist/TCP through a fixed, embodiment-specific $\mathrm{SE}(3)$ transform. Candidate bases are evaluated over the sequence using IK feasibility, joint limits, collision checks, and trajectory quality. A human operator selects one of the top-three valid sequence-level solutions, and the sequence is excluded if none is plausible. Appendix~\ref{app:data_curation_details} provides the search range, ranking criteria, and acceptance procedure.

Following kinematic alignment, we render the target URDF under the matched egocentric camera view and composite the rendered foreground into the recovered scene. This initial composite is then passed through the harmonization post-processing described in Section~\ref{sec:harmonization_postprocess}. The resulting harmonized composite constitutes the final pseudo-GT reference.

Automatic checks and human screening are applied to intermediate outputs from segmentation, background restoration, retargeting, IK and virtual-base placement, rendering/compositing, and harmonization. Samples that fail the stage-specific criteria are excluded, and the retained harmonized outputs constitute the pseudo-GT reference set used for similarity-based evaluation. Appendix~\ref{app:data_curation_details} provides the detailed checks and rejection criteria.

\subsection{HandEdit Dataset}
HandEdit dataset is built from the egocentric portions of five public hand-object datasets: EgoDex~\cite{egodex}, ARCTIC~\cite{arctic}, OakInk2~\cite{oakink2}, HOI4D~\cite{hoi4d}, and HO-Cap~\cite{hocap}. As shown in Table~\ref{tab:dataset_sources} and Appendix~\ref{app:source_datasets}, by combining these sources with cross-embodiment retargeting, rendering, and manual quality control, the current release contains over 200M image-level editing instances derived from 300K+ clips. It spans 600+ scenes, 400+ tasks, and 1,100+ objects.

Beyond benchmark evaluation, HandEdit provides paired human-source and robot-embodied images for training human-to-robot editing models. These models can directly convert human-hand manipulation frames into robot-centric observations, providing scalable data for policy pre-training and reducing the need for costly real-robot or teleoperated data collection. Prior work has shown that robot-centric data generated from human videos can support policy pre-training and co-training for real-robot manipulation~\cite{dexumi,masquerade,h2r,phantom,heng2025rwor,mitty,h2rgrounder}.

\subsection{HandEdit Benchmark}
HandEdit defines two human-to-robot dexterous hand editing tracks: \textbf{\texttt{Hand-only}} and \textbf{\texttt{Hand-Arm}}. The \textbf{\texttt{Hand-only}} track edits the local hand region, replacing the human hand with a dexterous robotic hand while preserving the hand-object interaction. The \textbf{\texttt{Hand-Arm}} track additionally edits the forearm or full arm, making the task more challenging because it requires larger-region background preservation, plausible arm kinematics, and consistent hand-object-scene composition. 

To instantiate these two tracks, we construct a diverse URDF collection with 13 standalone dexterous-hand URDFs and 13 hand-arm URDFs, covering representative hand families such as Ability, Allegro, DexHand(021), Inspire(RH56DFX and RH5DG2), Leap, OrcaHand, Revo2, Schunk SVH, Shadow Hand, Sharpa, Wuji, and RoHand, together with arm platforms including Jaka, KUKA, Panda, RM(65, 75), UR5, and xArm. Since many publicly available URDFs lack consistent material and color specifications, we standardize their visual assets by adding unified materials and colors, ensuring visually consistent rendering across different embodiments. The full URDF list is provided in Appendix~\ref{app:urdf_roster}. This diversity enables evaluation across substantially different hand morphologies, wrist structures, and arm geometries within a unified image-editing setting.

\begin{table}[!t]
\centering
\caption{\textbf{Overview of the Source Datasets in HandEdit.}}
\label{tab:dataset_sources}
\resizebox{\textwidth}{!}{
\begin{tabular}{c|ccc|ccc}
\toprule
\textbf{Dataset}  & \multicolumn{3}{c|}{\textbf{Data Statistics}} & \multicolumn{3}{c}{\textbf{Content Diversity}} \\
\cmidrule(lr){2-4}\cmidrule(l){5-7}
& \textbf{Frames} & \textbf{Clips} & \textbf{Resolution} & \textbf{Scenes} & \textbf{Objects} & \textbf{Tasks} \\
\midrule
EgoDex~\cite{egodex}  & 90M & 338K & 1080p & Tabletop & 500 Household Objects & 194 Tasks \\
ARCTIC~\cite{arctic}  & 2.1M & 339 & 2800 $\times$ 2000 & MoCap Studio & 11 Articulated Objects & Tool-use / Grasp \\
OakInk2~\cite{oakink2} & 4.01M & 627 & 848 $\times$ 480 & Tabletop & 75 Objects & 150 Tasks \\
HOI4D~\cite{hoi4d} & 2.4M & 4K & 1280 $\times$ 800 & Rooms & 800 Inst. / 16 Cats. & 54 Tasks \\
HO-Cap~\cite{hocap} & 656K & 64 & multi-res. & Tabletop & 64 Daily Objects & Pick-place / Handover / Tool-use \\
\midrule
\rowcolor{gray!15}
\textbf{HandEdit} & \textbf{200M} & \textbf{300K+} & \textbf{multi-res.} & \textbf{600+ Scenes} & \textbf{1.1K+ Objects} & \textbf{400+ Tasks} \\
\bottomrule
\end{tabular}
}
\end{table}

\subsection{Evaluation Metrics}
\label{sec:metric}
\begin{figure}[t!]
\centering
\includegraphics[width=0.95\textwidth]{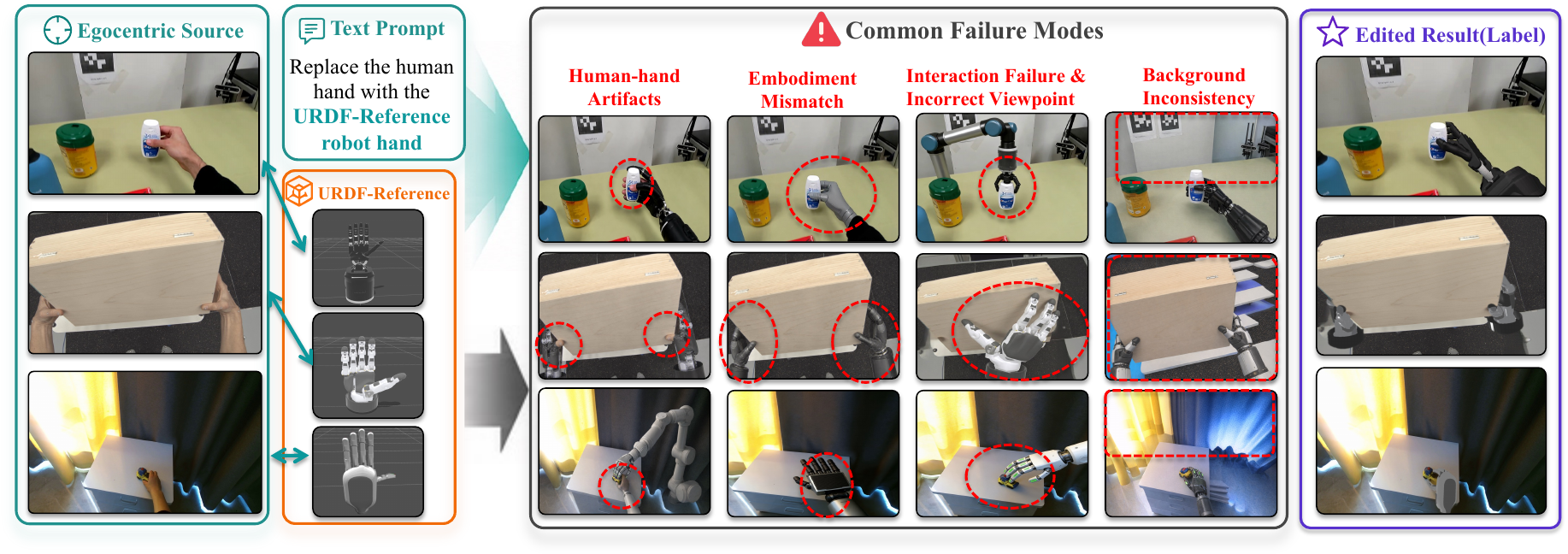}
\caption{\textbf{Common Failures of Existing Models in URDF-conditioned Hand.} Existing image editing models struggle to faithfully replace human hands with the target URDF-specified robot hand while preserving the original scene and interaction. Failures include residual human-hand artifacts, embodiment mismatch, interaction failure, incorrect viewpoint, and background inconsistency.}
\label{fig:failure_modes}
\end{figure}

To evaluate both visual fidelity and embodied identity correctness, HandEdit introduces a multi-dimensional metric suite, including \textbf{Generic Similarity Metrics}, \textbf{VLM-based Judgment}, and \textbf{Embodiment-Aware Metrics}. As illustrated in Figure~\ref{fig:failure_modes}, URDF-conditioned hand editing may suffer from diverse failure modes, such as residual human-hand artifacts, embodiment mismatch, incorrect viewpoint, interaction failure, and background inconsistency. Generic Similarity Metrics measure low-level and perceptual similarity between the edited image and the pseudo-GT target across the full image, edited ROI, and background, capturing overall visual fidelity and scene preservation quality. VLM-based Judgment and Embodiment-Aware Metrics further assess embodiment-specific correctness, including whether the human hand is properly removed, whether the generated robot hand matches the target robot, and whether the original interaction is consistently preserved. While VLM-based Judgment evaluates these aspects \textbf{\textit{at the semantic level}}, Embodiment-Aware Metrics assess them \textbf{\textit{at the geometric level}} by focusing on structural alignment, viewpoint consistency, and hand-object interaction preservation.


\textbf{Generic Similarity Metrics.}
Following prior image editing benchmarks~\cite{egoedit, motionedit, yu2024anyedit}, we compute generic similarity metrics over three regions: the full image, the edited ROI, and the background.
\begin{align}
\mathcal{M}_{\mathrm{region}}=\{\mathrm{PSNR},\mathrm{SSIM},\mathrm{LPIPS},\mathrm{FID}\}.
\end{align}
PSNR~\cite{hore2010image} measures pixel-level reconstruction fidelity, SSIM~\cite{wang2004image} evaluates structural similarity, LPIPS~\cite{zhang2018unreasonable} measures perceptual similarity using deep features, and FID~\cite{heusel2017gans} compares the distributional realism of generated images against reference images. When computed outside the edited ROI $M$, these metrics also quantify background consistency, i.e., how well the unchanged scene content is preserved. However, these metrics mainly capture low-level fidelity, perceptual similarity, or distributional realism, and ~\textbf{\textit{do not directly determine whether the edited result matches the target robot embodiment correctly or whether the original interaction has been preserved.}} Therefore, we introduce VLM-based judgment and embodiment-aware metrics to support comprehensive evaluation.

\begin{figure}[t!]
\centering
\includegraphics[width=0.9\textwidth]{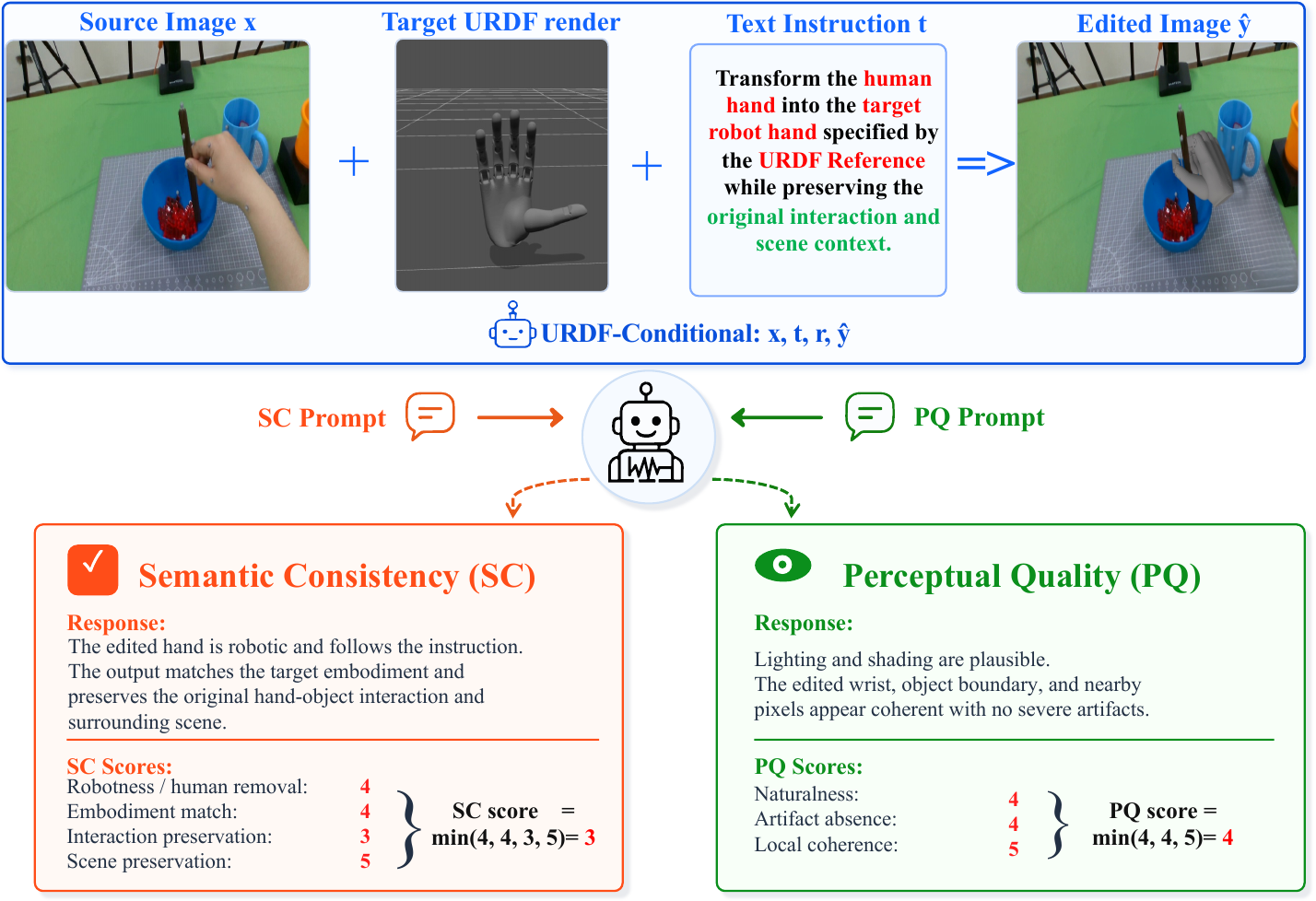}
\caption{\textbf{VLM-based Judgment.} Given source image, target URDF render, text instruction, and edited result, the VLM evaluates the edit from two complementary perspectives: Semantic Consistency (SC), measuring embodiment correctness, human-hand removal, interaction preservation, and scene consistency; and Perceptual Quality (PQ), which assesses identity naturalness, artifact absence, and local coherence. The final VLM score is computed by the normalized SC and PQ scores.}
\label{fig:vlm_metric}
\end{figure}

\textbf{VLM-based Judgment.}
We adopt a GPT-4o~\cite{openai2024gpt4ocard} judgment protocol following recent MLLM-based evaluation frameworks~\cite{viescore, lee2023holistic} for image synthesis and editing. In particular, as shown in Figure~\ref{fig:vlm_metric}, we decompose semantic-level evaluation into Semantic Consistency (SC) and Perceptual Quality (PQ). SC measures whether the edited result satisfies the requested robot embodiment and preserves the intended task conditions, whereas PQ evaluates whether the edited image appears visually natural and locally coherent. The final VLM score is defined as
\begin{align}
S_{\mathrm{vlm}}=\sqrt{SC\cdot PQ}.
\end{align}
Here both $SC$ and $PQ$ are normalized to $[0,1]$. This score serves as a semantic-level complement to generic similarity metrics and embodied metrics. The standardized judgment prompts are provided in Appendix~\ref{app:vlm_protocol}.

\textbf{Embodiment-Aware Metrics.}
We introduce embodiment-aware metrics, assessing three aspects: Human-hand Removal, Target-robot Fidelity, and Interaction consistency. We denote the source image as $x$, the pseudo-GT target as $y$, and the edited output as $\hat{y}$, and define the ROI mask $M = M_h \cup M_r$, where $M_h$ is the source human hand or hand-arm ROI and $M_r$ is the edited robot ROI.

\begin{itemize}[noitemsep,topsep=0pt,leftmargin=*,itemsep=2pt]
\item \textbf{Human-hand Removal.} 
We penalize residual human-hand artifacts within the edited region, such as exposed skin or incomplete finger removal. For this metric, we use the ROI mask $M$. Let $N_{\mathrm{skin}}(\hat{y};M)$ denote the number of skin-like pixels detected within $M$, and let $N(M)$ be the number of pixels in $M$:
\begin{align}
S_{\mathrm{rem}}=1-\frac{N_{\mathrm{skin}}(\hat{y};M)}{N(M)}.
\end{align}

\item \textbf{Target-robot Fidelity.} 
We evaluate the consistency between the edited robot hand and the target embodiment. Target-robot Fidelity is decomposed into two complementary components: structural fidelity, which measures geometric consistency, and ID fidelity, which assesses whether the generated robot hand matches the target robot embodiment.

\textbf{For structural fidelity}, we compare the edited ROI with the pseudo-GT target using a frozen DINOv2 encoder~\cite{dinov2}. Let $\phi_{\mathrm{str}}(\cdot)$ be the DINOv2 image encoder. The structural score  is defined as
\begin{align}
S_{\mathrm{struct}}
=\frac{1}{2}
\left(1+\cos\left(
\phi_{\mathrm{str}}(\hat{y}\odot M),
\phi_{\mathrm{str}}(y\odot M)
\right)
\right).
\end{align}

\textbf{For identity fidelity}, we compare the edited robot region with the URDF render bank of the requested embodiment, provided in Appendix~\ref{app:embodiment_description_bank}. 
Denote $\{(I^{\mathrm{ref}}_{v}, M^{\mathrm{ref}}_{v})\}_{v=1}^{V}$ as the rendered reference views of the requested embodiment, where $I^{\mathrm{ref}}_{v}$ is the $v$-th rendered image and $M^{\mathrm{ref}}_{v}$ is its robot-hand mask. We encode the edited robot ROI and the reference views with a frozen CLIP encoder~\cite{radford2021learning}, and define the URDF-reference consistency score by the maximum similarity over all rendered views:
\begin{align}
S_{\mathrm{ref}}
=
\max_{v}
\frac{1}{2}
\left(
1+
\cos\left(
\phi_{\mathrm{clip}}(\hat{y}\odot M_r),
\phi_{\mathrm{clip}}(I^{\mathrm{ref}}_{v}\odot M^{\mathrm{ref}}_{v})
\right)
\right).
\end{align}
Unlike the structural term, $S_{\mathrm{ref}}$ is not computed against the pseudo-GT image. It measures whether the edited region is closest to the requested robot embodiment. CLIP captures the overall robot identity but is less sensitive to local color assignments, such as dark fingertips, white shells, or colored joints. We therefore compare the edited robot ROI with the pseudo-GT target in CIELAB color space. Let $\Delta E$ be the average pixel-wise CIELAB distance over the overlapping foreground mask region $M_r$.We define
\begin{align}
S_{\mathrm{Lab}} &= \exp(-\Delta E / \tau_{\mathrm{Lab}}), \\
S_{\mathrm{ID}} &= \lambda S_{\mathrm{ref}} + (1-\lambda) S_{\mathrm{Lab}} .
\end{align}
We use $\lambda=0.5$ to balance URDF-reference consistency and local color consistency, and set $\tau_{\mathrm{Lab}}=25$ in all experiments.

\item \textbf{Interaction consistency.} Since the hand region is intentionally replaced by a robot embodiment, we do not measure interaction consistency inside the edited hand itself. Instead, we evaluate whether the manipulated object and its nearby contact neighborhood are preserved. We compute LPIPS on the object-contact region $M_O$, defined by dilating the object mask around the contact area:
\begin{align}
S_{\mathrm{int}}=1-\operatorname{LPIPS}(x\odot M_{O},\hat{y}\odot M_{O}).
\end{align}
A higher score indicates better preservation of object appearance and local interaction context.
\end{itemize}

\subsection{Harmonization Post-processing}
\label{sec:harmonization_postprocess}
The rendering-and-compositing stage produces an initial pseudo-GT image by placing the rendered robot foreground into the restored scene. Because the foreground and background are processed separately, the initial composite may contain mismatches in color, illumination, contrast, texture, or boundary appearance, making the inserted robot look visually detached from its surroundings. We therefore append a lightweight image-harmonization module to the end of the data curation pipeline.

We adopt Harmonizer~\cite{Harmonizer} ($\sim$20 MB) and train it on 10,000 natural egocentric hand images using a self-supervised procedure. Following the standard image-harmonization paradigm, we apply appearance augmentations to the masked hand or hand-arm region to synthesize an inharmonious input and use the original image as the reconstruction target. The augmentations cover lighting, color, contrast, and boundary appearance. At inference time, the rendered robot mask localizes refinement to the robot foreground and its boundary. The module adjusts the appearance of the composite without changing the robot joint configuration, wrist pose, object state, or interaction geometry. The harmonized outputs are used as the pseudo-GT references for all subsequent experiments.

\section{Experiments}
\subsection{Experimental Setup}
We construct the official HandEdit test set with two evaluation tracks: Hand-only and Hand-Arm. Each track contains 1K images paired with 13 corresponding target embodiments. The harmonized outputs of the complete data curation pipeline serve as the pseudo-GT references for all reported evaluations. All results are reported at the image level using the evaluation metrics introduced in Section~\ref{sec:metric}.

We benchmark 11 representative commercial and open-source image editors. The model details refer to Appendix~\ref{app:benchmark_model}. To ensure comparability, prompt templates are fixed within each benchmark track, and individual models are evaluated under conditioning interfaces they natively support. Commercial API-based models are evaluated within the same evaluation period using each platform's default inference configuration to reduce the effect of API-side model updates. Open-source models are evaluated using official checkpoints and standardized inference settings. The standardized instruction templates used for benchmark evaluation are provided in Appendix~\ref{app:benchmark_execution}.

\begin{table*}[t!]
\centering
\caption{\textbf{Hand-only Generic Similarity.} PSNR/SSIM/LPIPS are reported on full images, background regions, and ROI crops. FID is reported on full images and ROI crops. Best and second-best model scores are bolded and underlined, respectively.}
\label{tab:handonly_generic}
\scriptsize
\setlength{\tabcolsep}{4.5pt}
\renewcommand{\arraystretch}{1.08}
\resizebox{\textwidth}{!}{
\begin{tabular}{l|ccc|ccc|ccc|cc}
\toprule
\multirow{2}{*}{\textbf{Model}} & \multicolumn{3}{c|}{\textbf{PSNR}$\uparrow$} & \multicolumn{3}{c|}{\textbf{SSIM}$\uparrow$} & \multicolumn{3}{c|}{\textbf{LPIPS}$\downarrow$} & \multicolumn{2}{c}{\textbf{FID}$\downarrow$} \\
\cmidrule(lr){2-4}\cmidrule(lr){5-7}\cmidrule(lr){8-10}\cmidrule(l){11-12}
 & Full & BG & ROI & Full & BG & ROI & Full & BG & ROI & Full & ROI \\
\midrule
GPT-Image-2~\cite{openaiIntroducingChatGPT} & \textbf{19.85} & \textbf{22.82} & \underline{12.00} & 0.781 & 0.828 & 0.584 & \textbf{0.224} & \textbf{0.150} & \textbf{0.482} & \textbf{58.83} & \textbf{99.71} \\
Nano-Banana-2~\cite{geminiNanoBanana} & \underline{18.92} & \underline{21.92} & 11.12 & \textbf{0.814} & \textbf{0.865} & \underline{0.594} & \underline{0.228} & \underline{0.152} & \underline{0.501} & \underline{74.77} & \underline{103.83} \\
GPT-Image-1.5~\cite{openaiChatGPTImages} & 15.68 & 16.26 & \textbf{12.13} & 0.678 & 0.722 & 0.542 & 0.466 & 0.393 & 0.611 & 89.60 & 123.14 \\
Seedream-4.5~\cite{seedream2025seedream} & 18.31 & 20.98 & 11.67 & \underline{0.792} & \underline{0.839} & \textbf{0.595} & 0.296 & 0.222 & 0.537 & 79.15 & 117.90 \\
Flux-2-Pro~\cite{flux-2-2025} & 13.13 & 13.37 & 11.41 & 0.606 & 0.655 & 0.480 & 0.590 & 0.519 & 0.683 & 99.90 & 108.36 \\
Hunyuan-Image-3.0~\cite{cao2025hunyuanimage} & 13.37 & 13.90 & 10.81 & 0.631 & 0.678 & 0.513 & 0.565 & 0.490 & 0.667 & 102.19 & 113.40 \\
Nano-Banana~\cite{nanobananaNanoBanana} & 13.17 & 13.38 & 11.79 & 0.627 & 0.670 & 0.527 & 0.624 & 0.548 & 0.713 & 99.57 & 117.97 \\
Qwen-Image-Edit-2511~\cite{wu2025qwenimage} & 12.14 & 12.25 & 11.43 & 0.593 & 0.638 & 0.468 & 0.659 & 0.588 & 0.737 & 129.99 & 142.18 \\
Flux-Kontext-Max~\cite{labs2025flux} & 12.08 & 12.23 & 11.28 & 0.594 & 0.639 & 0.488 & 0.639 & 0.565 & 0.720 & 127.89 & 136.10 \\
Omnigen2~\cite{wu2025omnigen2} & 12.60 & 12.75 & 11.70 & 0.622 & 0.665 & 0.512 & 0.632 & 0.555 & 0.752 & 98.90 & 127.39 \\
FireRed-Image-Edit-1.1~\cite{firered2026rededit} & 12.43 & 12.54 & 11.84 & 0.605 & 0.647 & 0.505 & 0.650 & 0.576 & 0.728 & 149.12 & 144.16 \\
\bottomrule
\end{tabular}}
\end{table*}

\begin{table*}[t]
\centering
\caption{\textbf{Hand-only Embodiment-Aware Metrics and VLM Judgment.} Best and second-best model scores are bolded and underlined, respectively.}
\label{tab:handonly_embodied_vlm}
\scriptsize
\setlength{\tabcolsep}{4.8pt}
\renewcommand{\arraystretch}{1.08}
\resizebox{\textwidth}{!}{
\begin{tabular}{l|cccc|ccc}
\toprule
\multirow{2}{*}{\textbf{Model}} & \multicolumn{4}{c|}{\textbf{Embodied Task Metrics}$\uparrow$} & \multicolumn{3}{c}{\textbf{VLM Judge}$\uparrow$} \\
\cmidrule(lr){2-5}\cmidrule(l){6-8}
 & Removal & Struct Fidelity & ID Fidelity & Interaction & SC & PQ & VLM \\
\midrule
Gpt-Image-2~\cite{openaiIntroducingChatGPT} & 0.953 & \textbf{0.780} & 0.852 & \textbf{0.703} & \underline{0.785} & 0.590 & 0.663 \\
Nano-Banana-2~\cite{geminiNanoBanana} & 0.925 & \underline{0.746} & 0.792 & \underline{0.617} & 0.762 & 0.669 & \underline{0.692} \\
Gpt-Image-1.5~\cite{openaiChatGPTImages} & \textbf{0.955} & 0.730 & \textbf{0.915} & 0.435 & \textbf{0.862} & \underline{0.698} & \textbf{0.765} \\
Seedream-4.5~\cite{seedream2025seedream} & 0.931 & 0.723 & 0.758 & 0.608 & 0.568 & 0.662 & 0.570 \\
Flux-2-Pro~\cite{flux-2-2025} & 0.939 & 0.723 & 0.759 & 0.375 & 0.729 & 0.654 & 0.672 \\
Hunyuan-Image-3.0~\cite{cao2025hunyuanimage} & 0.933 & 0.716 & 0.851 & 0.389 & 0.762 & 0.602 & 0.662 \\
Nano-Banana~\cite{nanobananaNanoBanana} & 0.923 & 0.693 & 0.753 & 0.329 & 0.463 & 0.521 & 0.432 \\
Qwen-Image-Edit~\cite{wu2025qwenimage} & \underline{0.954} & 0.670 & 0.619 & 0.306 & 0.435 & 0.694 & 0.472 \\
Flux-Kontext-Max~\cite{labs2025flux} & 0.952 & 0.660 & 0.746 & 0.274 & 0.357 & 0.598 & 0.363 \\
Omnigen2~\cite{wu2025omnigen2} & 0.904 & 0.678 & 0.529 & 0.266 & 0.538 & 0.512 & 0.471 \\
FireRed-Image-Edit-1.1~\cite{firered2026rededit} & 0.952 & 0.630 & \underline{0.890} & 0.293 & 0.228 & \textbf{0.754} & 0.292 \\
\rowcolor{gray!20} \textcolor{gray!50}{Pseudo-GT}  & \textcolor{gray!50}{0.964} & \textcolor{gray!50}{1.000} & \textcolor{gray!50}{0.964} & \textcolor{gray!50}{0.821} & \textcolor{gray!50}{0.776} & \textcolor{gray!50}{0.492} & \textcolor{gray!50}{0.623} \\
\bottomrule
\end{tabular}}
\end{table*}

\begin{table*}[t]
\centering
\caption{\textbf{Hand-Arm Generic Similarity.} PSNR/SSIM/LPIPS are reported on full images, background regions, and ROI crops. FID is reported on full images and ROI crops. Best and second-best model scores are bolded and underlined, respectively.}
\label{tab:handarm_generic}
\scriptsize
\setlength{\tabcolsep}{4.5pt}
\renewcommand{\arraystretch}{1.08}
\resizebox{\textwidth}{!}{
\begin{tabular}{l|ccc|ccc|ccc|cc}
\toprule
\multirow{2}{*}{\textbf{Model}} & \multicolumn{3}{c|}{\textbf{PSNR}$\uparrow$} & \multicolumn{3}{c|}{\textbf{SSIM}$\uparrow$} & \multicolumn{3}{c|}{\textbf{LPIPS}$\downarrow$} & \multicolumn{2}{c}{\textbf{FID}$\downarrow$} \\
\cmidrule(lr){2-4}\cmidrule(lr){5-7}\cmidrule(lr){8-10}\cmidrule(l){11-12}
 & Full & BG & ROI & Full & BG & ROI & Full & BG & ROI & Full & ROI \\
\midrule
GPT-Image-2~\cite{openaiIntroducingChatGPT} & \textbf{15.07} & \underline{18.70} & 9.59 & 0.748 & 0.843 & \underline{0.592} & \underline{0.301} & \underline{0.164} & \textbf{0.514} & \textbf{108.90} & \textbf{146.07} \\
GPT-Image-1.5~\cite{openaiChatGPTImages} & 13.87 & 15.40 & \textbf{10.09} & 0.679 & 0.771 & 0.567 & 0.478 & 0.342 & 0.592 & 127.99 & \underline{152.34} \\
Nano-Banana-2~\cite{geminiNanoBanana} & \underline{14.95} & \textbf{19.73} & 9.22 & \textbf{0.764} & \textbf{0.861} & \textbf{0.598} & \textbf{0.292} & \textbf{0.148} & \underline{0.515} & \underline{124.71} & 158.15 \\
Flux-2-Pro~\cite{flux-2-2025} & 14.60 & 17.65 & 9.47 & \underline{0.752} & \underline{0.848} & 0.586 & 0.341 & 0.199 & 0.555 & 128.23 & 157.08 \\
HunyuanImage-3.0~\cite{cao2025hunyuanimage} & 12.41 & 13.37 & 9.73 & 0.632 & 0.726 & 0.529 & 0.564 & 0.421 & 0.669 & 136.74 & 161.39 \\
Qwen-Image-Edit~\cite{wu2025qwenimage} & 12.81 & 14.67 & 8.71 & 0.673 & 0.771 & 0.541 & 0.460 & 0.319 & 0.599 & 147.08 & 165.79 \\
Seedream-4.5~\cite{seedream2025seedream} & 13.04 & 14.46 & 9.51 & 0.668 & 0.759 & 0.561 & 0.508 & 0.365 & 0.624 & 146.33 & 167.19 \\
Nano-Banana~\cite{nanobananaNanoBanana} & 11.23 & 11.79 & 9.35 & 0.598 & 0.689 & 0.521 & 0.634 & 0.491 & 0.700 & 166.07 & 164.08 \\
FireRed-Image-Edit-1.1~\cite{firered2026rededit} & 11.82 & 12.78 & 9.06 & 0.644 & 0.732 & 0.559 & 0.595 & 0.440 & 0.703 & 158.60 & 190.08 \\
Omnigen2~\cite{wu2025omnigen2} & 11.97 & 12.63 & \underline{9.80} & 0.630 & 0.716 & 0.539 & 0.635 & 0.484 & 0.741 & 142.16 & 178.56 \\
Flux-Kontext-Max~\cite{labs2025flux} & 11.25 & 11.74 & 9.71 & 0.594 & 0.681 & 0.528 & 0.673 & 0.530 & 0.727 & 189.22 & 199.52 \\
\bottomrule
\end{tabular}}
\end{table*}

\begin{table*}[t]
\centering
\caption{\textbf{Hand-Arm Embodiment-Aware Metrics and VLM Judgment.} Best and second-best model scores are bolded and underlined, respectively.}
\label{tab:handarm_embodied_vlm}
\scriptsize
\setlength{\tabcolsep}{4.8pt}
\renewcommand{\arraystretch}{1.08}
\resizebox{\textwidth}{!}{
\begin{tabular}{l|cccc|ccc}
\toprule
\multirow{2}{*}{\textbf{Model}} & \multicolumn{4}{c|}{\textbf{Embodied task metrics}$\uparrow$} & \multicolumn{3}{c}{\textbf{VLM judge}$\uparrow$} \\
\cmidrule(lr){2-5}\cmidrule(l){6-8}
 & Removal & Struct Fidelity & ID Fidelity & Interaction & SC & PQ & VLM \\
\midrule
Gpt-Image-2~\cite{openaiIntroducingChatGPT} & \underline{0.983} & \textbf{0.778} & \underline{0.563} & \underline{0.595} & \underline{0.888} & 0.716 & 0.786 \\
Gpt-Image-1.5~\cite{openaiChatGPTImages} & \textbf{0.985} & \underline{0.744} & \textbf{0.606} & 0.421 & \textbf{0.909} & \textbf{0.814} & \textbf{0.856} \\
Nano-Banana-2~\cite{geminiNanoBanana} & 0.968 & 0.726 & 0.503 & \textbf{0.634} & 0.691 & 0.756 & 0.673 \\
Flux-2-Pro~\cite{flux-2-2025} & 0.960 & 0.743 & 0.516 & 0.541 & 0.726 & 0.741 & 0.704 \\
Hunyuan-Image-3.0~\cite{cao2025hunyuanimage} & 0.977 & \underline{0.744} & 0.519 & 0.336 & 0.863 & 0.778 & \underline{0.812} \\
Qwen-Image-Edit~\cite{wu2025qwenimage} & 0.982 & 0.743 & 0.468 & 0.489 & 0.705 & \underline{0.798} & 0.705 \\
Seedream-4.5~\cite{seedream2025seedream} & 0.965 & 0.702 & 0.502 & 0.424 & 0.484 & 0.734 & 0.530 \\
Nano-Banana~\cite{nanobananaNanoBanana} & 0.978 & 0.706 & 0.471 & 0.275 & 0.856 & 0.786 & 0.808 \\
FireRed-Image-Edit-1.1~\cite{firered2026rededit} & \underline{0.983} & 0.707 & 0.510 & 0.270 & 0.349 & 0.635 & 0.400 \\
Omnigen2~\cite{wu2025omnigen2} & 0.912 & 0.707 & 0.417 & 0.236 & 0.467 & 0.463 & 0.380 \\
Flux-Kontext-Max~\cite{labs2025flux} & \underline{0.983} & 0.679 & 0.470 & 0.246 & 0.388 & 0.531 & 0.351 \\
\rowcolor{gray!20} \textcolor{gray!50}{Pseudo-GT} & \textcolor{gray!50}{0.963} & \textcolor{gray!50}{1.000} & \textcolor{gray!50}{0.962} & \textcolor{gray!50}{0.788} & \textcolor{gray!50}{0.782} & \textcolor{gray!50}{0.482} & \textcolor{gray!50}{0.644} \\
\bottomrule
\end{tabular}}
\end{table*}

\subsection{Benchmark Results and In-depth Analysis}
Tables~\ref{tab:handonly_generic} and~\ref{tab:handonly_embodied_vlm} report results on the Hand-only track, while Tables~\ref{tab:handarm_generic} and~\ref{tab:handarm_embodied_vlm} report results on the Hand-Arm track. We also conduct a blinded human evaluation on the test set, with details provided in Appendix~\ref{app:human_eval}. We summarize the key observations below.

\textbf{GPT-Image-2 provides the strongest overall baseline.}
GPT-Image-2 achieves consistently strong performance across both generic similarity metrics and embodiment-aware metrics. In particular, it obtains the best LPIPS(ROI) and FID(ROI) scores, suggesting better local editing quality in the manipulated region. It also achieves the highest structural fidelity and interaction scores, indicating stronger capability in matching the target robot embodiment and preserving hand-object interactions.

\textbf{VLM-based judgment is useful but not sufficient.}
Although GPT-Image-1.5 obtains the highest VLM score, it does not achieve the best performance on structural fidelity or interaction preservation. Conversely, GPT-Image-2 performs better on these embodiment-aware metrics. This discrepancy suggests that VLM-based judgment mainly captures high-level semantic consistency and perceptual quality, while fine-grained robot morphology and contact-level interaction are better reflected by embodiment-aware metrics.

\textbf{Perceptual quality alone does not guarantee editing task success.}
FireRed-Image-Edit-1.1 achieves the highest PQ score, indicating that its outputs are often visually plausible. However, its low SC and embodiment-aware scores show that visually natural edits may still fail to satisfy the target embodiment or preserve the intended interaction. This highlights the limitation of evaluating embodied image editing solely through perceptual realism.

\textbf{The main challenge lies in embodiment-aware editing.}
Most models obtain relatively high human-hand removal scores, suggesting that removing the source human hand is comparatively easier. In contrast, structural fidelity, ID fidelity, and interaction preservation show larger performance gaps across models. This indicates that the benchmark primarily evaluates whether models can generate the correct robot embodiment and maintain physically plausible interaction with objects.

\section{Conclusion}
In this work, we introduce HandEdit, the first dedicated dataset and benchmark for human-to-robot dexterous hand image editing. HandEdit provides a large-scale dataset comprising 200M editing samples across 26 embodiments, and introduces two evaluation tracks: Hand-only and Hand-Arm. The benchmark focuses on editing human hand-object images into robot-embodied hand-object images that are visually plausible and structurally useful for embodied learning. To address the limitations of existing image-editing evaluations, HandEdit further introduces a multi-dimensional evaluation suite that integrates generic similarity metrics, VLM-based judgment, and embodiment-aware metrics. Finally, we benchmark 11 representative commercial and open-source image editors, providing a comprehensive empirical analysis of current model capabilities and limitations in human-to-robot dexterous hand editing. Beyond evaluation, the paired data can also support human-to-robot editing models that generate robot-centric visual observations from human videos for downstream policy pre-training.  We hope that HandEdit will serve as a useful resource for both the embodied AI and image editing communities, facilitating future research on robot-centric data generation, dexterous embodiment transfer, and controllable image editing. 


\clearpage
\bibliographystyle{unsrtnat}
\bibliography{bibliography_short,references}

\clearpage
\bookmarksetup{startatroot}
\phantomsection
\pdfbookmark[1]{Supplementary Materials}{supplementary-materials}
{\noindent\Large\itshape Supplementary Materials\par}

\setcounter{section}{0}
\setcounter{subsection}{0}
\setcounter{figure}{0}
\setcounter{table}{0}

\renewcommand{\thesection}{\Alph{section}}
\renewcommand{\thesubsection}{\Alph{section}.\arabic{subsection}}
\renewcommand{\thefigure}{S\arabic{figure}}
\renewcommand{\thetable}{S\arabic{table}}

\renewcommand{\theHsection}{supp.\Alph{section}}
\renewcommand{\theHsubsection}{supp.\Alph{section}.\arabic{subsection}}
\renewcommand{\theHfigure}{suppfig.\arabic{figure}}
\renewcommand{\theHtable}{supptab.\arabic{table}}

\section{Dataset Resources and Benchmark Assets}\label{app:dataset_resources}

\subsection{Full Target URDF Roster}\label{app:urdf_roster}

HandEdit covers 26 target embodiments, including 13 hand-only embodiments and 13 hand-arm embodiments. The roster is designed to span substantial variation in hand morphology, wrist structure, arm geometry, and embodiment scale. The hand-only track emphasizes localized dexterous-hand embodiment editing, whereas the hand-arm track introduces larger edited regions and stronger requirements on scene preservation, pose continuity, and arm-hand consistency. 

Note that we construct the target embodiment set from publicly available robot URDF assets. Part of the hand models are adapted from DexSuite's~\cite{bunny-visionpro} \texttt{dex-urdf} repository, which provides high-quality dexterous hand and object models in URDF format and is released under the MIT License. The remaining embodiments are collected from the corresponding official open-source robot releases. All URDF assets are used for non-commercial scientific research and benchmark evaluation only.

\begin{table}[H]
\centering
\small
\caption{Full target URDF roster used in HandEdit.}
\label{tab:appendix_urdf_roster}
\begin{tabularx}{\textwidth}{@{}L{2.2cm}X@{}}
\toprule
\textbf{Category} & \textbf{URDFs} \\
\midrule
Hand-only & Allegro, Revo2, DexHand021, Leap, Orca, RH56DFX, RH5DG2, RoHand, Schunk Hand, Shadow Hand, Sharpa, Wuji \\
Hand-arm & Jaka Zu7 + DexHand021, KUKA + Sharpa, Panda + Allegro, Panda + Orca, RM65 + BrainCo, RM75 + RoHand, UR5 + RH56DFX, UR5 + RH5DG2, UR5 + Schunk Hand, UR5 + Shadow Hand, UR5 + Wuji, xArm + Ability, xArm + Leap \\
\bottomrule
\end{tabularx}
\end{table}

\subsection{Source Datasets}\label{app:source_datasets}
HandEdit Dataset is constructed from the egocentric portions of five public hand-object datasets: EgoDex, ARCTIC, OakInk2, HOI4D, and HO-Cap. These sources are complementary in scale, scene types, interaction patterns, and capture quality, and together provide broad coverage of dexterous hand-object interactions across both in-the-wild daily-life scenes and more controlled laboratory-style capture settings.

\begin{figure}[H]
\centering
\includegraphics[width=1\textwidth,height=2.1in,keepaspectratio]{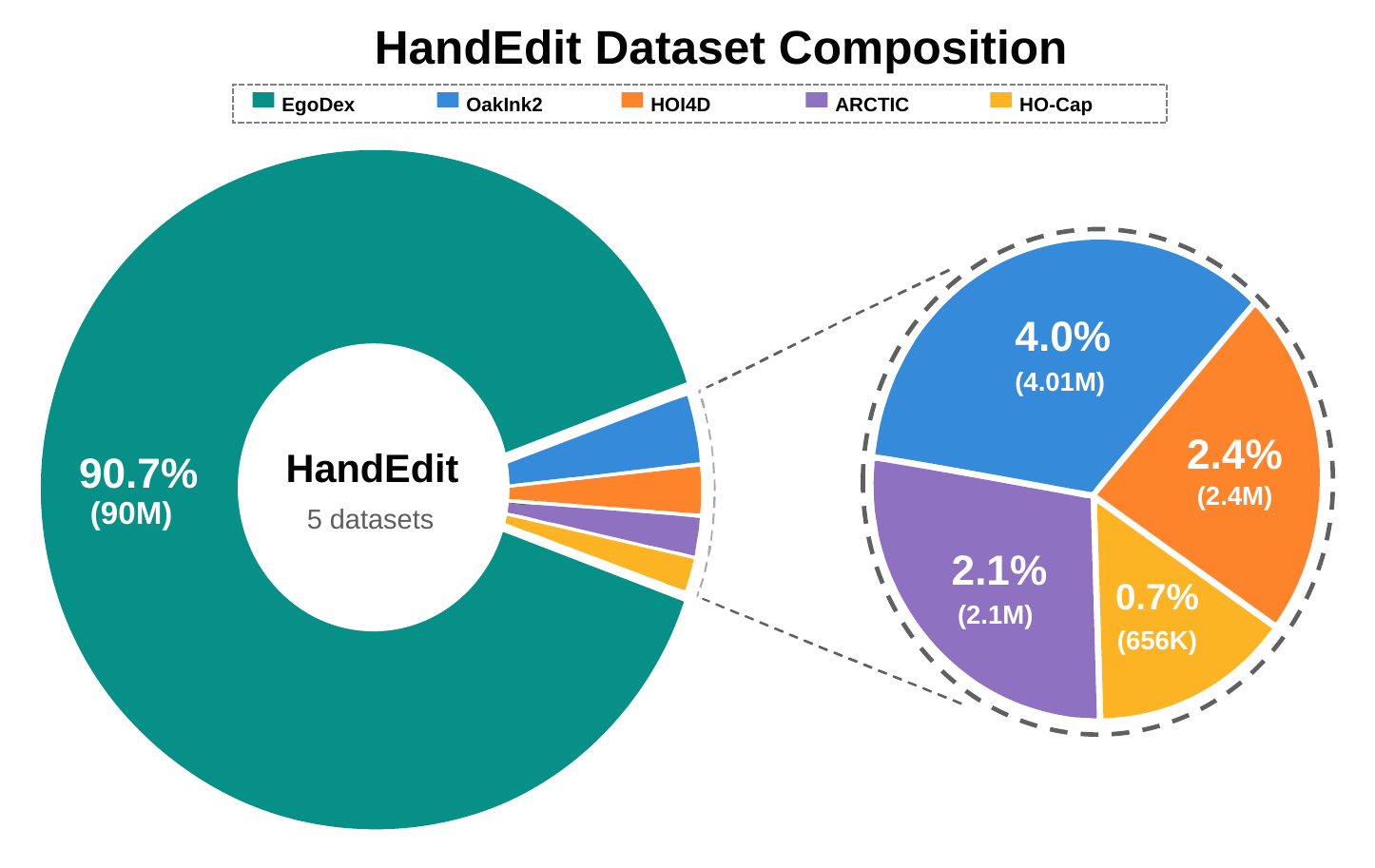}
\caption{\textbf{Data Composition of HandEdit.}}
\label{fig:data_composition}
\end{figure}

\paragraph{EgoDex.}
EgoDex provides the largest source pool in HandEdit. It contains approximately 90 million frames and 338K task-demonstration clips. The dataset focuses on tabletop environments and covers 194 daily household manipulation tasks with roughly 500 household objects, contributing large amounts of first-person interaction footage with varied grasp types, contact patterns, and dexterous behaviors. The dataset is available under CC-by-NC-ND terms.

\paragraph{ARCTIC.}
ARCTIC is designed for dexterous bimanual manipulation and dynamic contact. It contains about 2.1 million frames across 339 sequences, recorded from 10 subjects using eight third-person views and one egocentric view, and centers on interaction with 11 articulated objects. Its interactions are organized around use and grasp intents, making ARCTIC particularly valuable for evaluating whether edited robot hands preserve contact structure under articulated-object interaction. The data and software are released under a copyright license that permits use for non-commercial scientific research purposes only.

\paragraph{OakInk2.}
OakInk2 targets complex interaction understanding from the perspective of object affordances. It provides about 4.01 million frames and 627 manipulation sequences, covering 75 objects, four manipulation scenarios, and 150 complex tasks. Compared with the other sources, OakInk2 places stronger emphasis on long-horizon, multi-stage, and often bimanual manipulation.

\paragraph{HOI4D.}
HOI4D contributes dense egocentric RGB-D hand-object interaction data in diverse indoor environments. It contains roughly 2.4 million frames and around 4,000 sequences, spanning 610 indoor rooms, 800 object instances from 16 object categories, and 54 functionality-oriented interaction tasks. The dataset is available under the MIT license.

\paragraph{HO-Cap.}
HO-Cap emphasizes accurate 3D hand-object pose capture and reconstruction using a low-cost multi-view RGB-D capture setup. It contains 656K frames over 64 sequences and covers 64 daily objects. Although smaller in scale, it provides pose-rich interaction data across both one-handed and two-handed manipulation scenarios, including pick-and-place, handover, and affordance-driven object use.
The dataset is available under the GPL-3.0 license.

\begin{figure*}[t]
\centering
\includegraphics[width=0.96\textwidth]{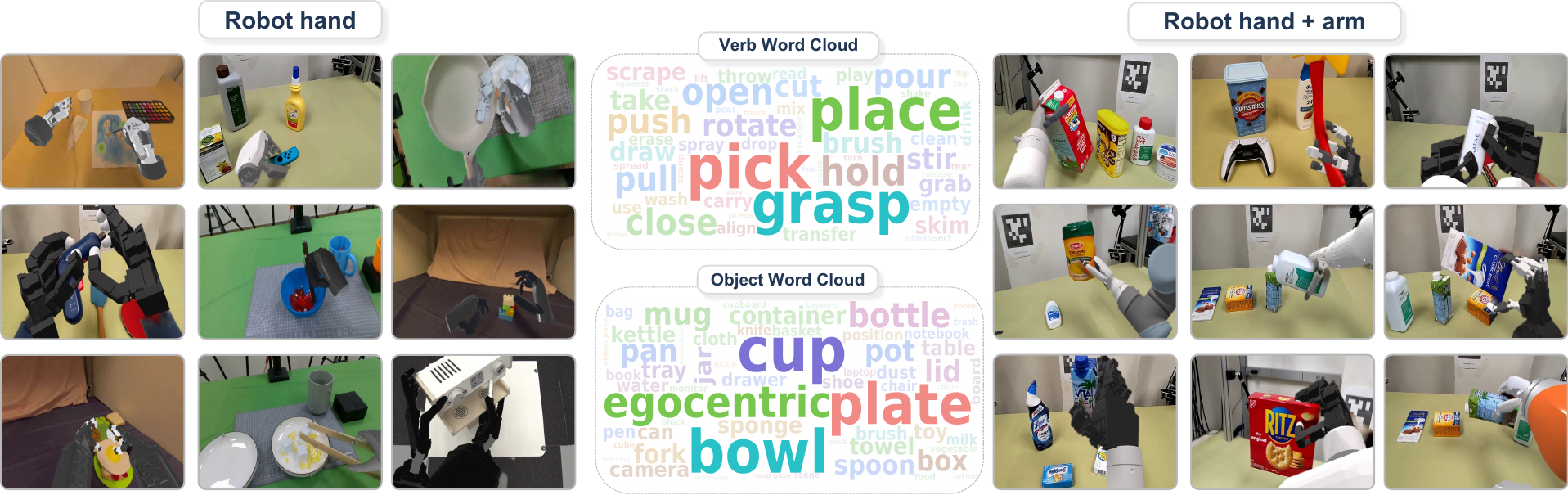}
\caption{\textbf{Word cloud visualization.}}
\label{fig:appendix_word_cloud}
\end{figure*}

\section{Data Curation Details}
\label{app:data_curation_details}

\subsection{Quality Control}

Each pipeline stage is checked before its output is passed to the next stage. Table~\ref{tab:stagewise_qc} lists the automatic checks, manual checks, and rejection rules used during pseudo-GT generation.

\begin{table}[H]
\centering
\scriptsize
\caption{Quality-control criteria for pseudo-GT generation. A dash indicates a manual-only check.}
\label{tab:stagewise_qc}
\setlength{\tabcolsep}{3.5pt}
\renewcommand{\arraystretch}{1.15}
\begin{tabularx}{\textwidth}{@{}L{2.15cm}XXX@{}}
\toprule
\textbf{Stage} & \textbf{Automatic checks} & \textbf{Manual checks} & \textbf{Reject if} \\
\midrule
Segmentation
& Non-empty mask; SAM 3 confidence $\geq 0.75$.
& Mask completeness; boundary accuracy; object preservation.
& Missing regions; object removal; severe under- or over-segmentation. \\

Background restoration / inpainting
& --
& Residue; ghosting; texture and object integrity.
& Visible residue; severe ghosting; structural corruption. \\

Hand retargeting
& Fingertip/wrist error; contact distance; joint-limit and penetration checks.
& Grasp/contact/release states; finger--object configuration.
& Fingertip/wrist error $>50$ mm; contact distance $>20$ mm; contact loss; incorrect interaction state; severe penetration. \\

IK and virtual-base placement
& Critical-frame IK feasibility; URDF joint limits; self-, support-plane-, and dual-arm collisions.
& Egocentric and third-person replay; collision; overextension; base plausibility.
& Critical-frame IK failure; joint-limit violation; collision; trajectory discontinuity; all top-three candidates implausible. \\

Rendering, compositing, and harmonization
& --
& Robot identity and structure; boundary; occlusion; scale; contact placement.
& Broken structure; incorrect occlusion; object corruption; contact misalignment. \\
\bottomrule
\end{tabularx}
\end{table}

The final harmonized composites undergo structure, object-integrity, occlusion, and contact-placement checks before being retained as pseudo-GT references.

\subsection{Failure Breakdown}

To quantify the sources of pseudo-GT rejection, we uniformly sampled 5,000 frames from the 734,864 ARCTIC frames marked as final non-kept examples. For each sampled frame, the annotator inspected the source image, human-mask overlay, restored background, pseudo-GT, and robot-mask overlay, and assigned one dominant primary failure cause. The processing log records an overall non-kept rate of approximately 33\% for this ARCTIC run. Because the candidate unit used by the processing pipeline is not identical to a raw source frame, we report the per-cause proportions within the sampled non-kept pool and do not extrapolate them to all raw frames.

\begin{table}[H]
\centering
\small
\caption{Primary causes among 5,000 uniformly sampled non-kept ARCTIC frames. Percentages are normalized within the sampled non-kept pool.}
\label{tab:pseudogt_failure_breakdown}
\begin{tabular}{lc}
\toprule
\textbf{Primary failure cause} & \textbf{Share of sampled non-kept frames} \\
\midrule
Hand retargeting & 3,173 / 5,000 (63.46\%) \\
Segmentation & 927 / 5,000 (18.54\%) \\
Background restoration / inpainting & 586 / 5,000 (11.72\%) \\
Rendering / compositing & 314 / 5,000 (6.28\%) \\
\midrule
Total & 5,000 / 5,000 (100\%) \\
\bottomrule
\end{tabular}
\end{table}

Retargeting is the dominant primary cause of rejection, followed by segmentation, background restoration, and rendering/compositing. The four-way annotation taxonomy aggregates IK and virtual-base failures within the reported categories and does not provide a separate IK-specific rate. Table~\ref{tab:pseudogt_failure_breakdown} describes rejected ARCTIC frames and does not measure residual errors among retained samples or failure distributions in the other source datasets.

\subsection{Virtual-Base Search and IK}

The virtual-base procedure separates automatic candidate generation and feasibility filtering from a single sequence-level human choice. For each source-sequence/target-embodiment pair, the human wrist is mapped to the robot wrist/TCP through a fixed, embodiment-specific $\mathrm{SE}(3)$ transform, and a selected candidate base remains fixed across all frames.

When the left and right shoulder keypoints are available, their midpoint, shoulder direction, and inter-shoulder distance are used to initialize the local base-search region. Otherwise, the wrist-trajectory center and the nominal reach of the target arm are used for initialization. Let $L_r$ denote the nominal reach of the target arm obtained from its URDF kinematic chain. To account for differences in arm scale and workspace across embodiments, the local grid varies lateral translation over $\{-0.10L_r,0,0.10L_r\}$, forward/backward translation over $\{-0.15L_r,0,0.15L_r\}$, and yaw over $\{-15^{\circ},0,15^{\circ}\}$, yielding $3\times3\times3=27$ candidates. Base height, roll, and pitch are fixed under the upright mounting assumption. Each candidate base remains fixed while it is evaluated over the full sequence. Candidates with IK-infeasible critical frames, joint-limit violations, or self-, support-plane-, or dual-arm collisions are discarded. Survivors are ranked lexicographically by critical-frame success, reachability, end-effector error, joint margin, and continuity.

For the top-three candidates, we render both egocentric and third-person replay videos. A human selects one feasible sequence-level candidate without modifying joint states, frame-level base poses, or search parameters. If all three candidates are implausible, the sequence is dropped. Candidate generation and feasibility checks are automatic, and human input is limited to the sequence-level selection.

\section{VLM-Based Judgment Protocol}\label{app:vlm_protocol}

In the main paper, HandEdit adopts a GPT-4o-based judgment protocol as a semantic-level complement to generic similarity metrics and embodied task metrics. The judge produces two scores: Semantic Consistency (SC) and Perceptual Quality (PQ).

\subsection{Score Definitions}

\paragraph{Semantic Consistency (SC).}
SC measures whether the edit satisfies the requested robot embodiment while preserving the task-relevant structure of the source interaction. We decompose it into four 1--5 sub-scores: \emph{robotness}, \emph{target-embodiment match}, \emph{interaction preservation}, and \emph{scene preservation}.

\paragraph{Perceptual Quality (PQ).}
PQ measures whether the edited result is visually coherent as a single image. It focuses on local plausibility around the edited region, including boundary naturalness, absence of visible artifacts, consistent geometry, and coherent lighting and occlusion. We decompose it into three 1--5 sub-scores: \emph{naturalness}, \emph{artifact absence}, and \emph{local coherence}.

Let $\alpha_{\mathrm{rob}}$, $\alpha_{\mathrm{id}}$, $\alpha_{\mathrm{int}}$, and $\alpha_{\mathrm{scene}}$ denote the SC sub-scores for robotness, target-embodiment match, interaction preservation, and scene preservation. Let $\beta_{\mathrm{nat}}$, $\beta_{\mathrm{art}}$, and $\beta_{\mathrm{coh}}$ denote the PQ sub-scores for naturalness, artifact absence, and local coherence. Following the conservative aggregation principle used in instruction-guided VLM evaluation~\citep{viescore}, we take the minimum sub-score within each aspect and then normalize the resulting scores to $[0,1]$:
\begin{align}
s_{\mathrm{SC}} &= \min(\alpha_{\mathrm{rob}}, \alpha_{\mathrm{id}}, \alpha_{\mathrm{int}}, \alpha_{\mathrm{scene}}) \\
s_{\mathrm{PQ}} &= \min(\beta_{\mathrm{nat}}, \beta_{\mathrm{art}}, \beta_{\mathrm{coh}}) \\
\hat{s}_{\mathrm{SC}} &= \frac{s_{\mathrm{SC}}-1}{4} \\
\hat{s}_{\mathrm{PQ}} &= \frac{s_{\mathrm{PQ}}-1}{4} \\
s_{\mathrm{vlm}} &= \sqrt{\hat{s}_{\mathrm{SC}}\ * \hat{s}_{\mathrm{PQ}}} 
\end{align}

\subsection{Semantic-Consistency Prompt Template}

\begin{table}[H]
\centering
\caption{Semantic-Consistency prompt template used for VLM-based judgment. }
\label{tab:appendix_task_prompt}
\setlength{\fboxsep}{7pt}
\fcolorbox{gray!35}{gray!20}{%
\begin{minipage}{1\textwidth}
\textbf{Semantic-Consistency Prompt Template}

\vspace{2pt}
You are evaluating an image editing result for a benchmark on egocentric hand-to-dexterous-robot image editing.

\vspace{4pt}
\textbf{You are given:}

\vspace{2pt}
\begin{tabularx}{\linewidth}{@{}p{0.04\linewidth}X@{}}
(1) & a source image showing a human hand or hand-arm interacting with an object;\\
(2) & an edited image;\\
(3) & a text instruction describing the target robot embodiment;\\
(4) & a URDF reference render of the target robot embodiment.
\end{tabularx}

\vspace{4pt}
\textbf{Your task} is to judge whether the edited image successfully transforms the human hand or hand-arm region into the requested dexterous robot embodiment while preserving the original object interaction and surrounding scene structure.

\vspace{4pt}
Please assign four 1--5 sub-scores:

\vspace{2pt}
\begin{tabularx}{\linewidth}{@{}p{0.31\linewidth}X@{}}
\textbf{Robotness} &
Whether the edited region is clearly robotic rather than human-like.\\
\textbf{Target-embodiment match} &
Whether the robot matches the requested target embodiment, using the text instruction and URDF reference render as cues.\\
\textbf{Interaction preservation} &
Whether the original object state, contact relationship, and hand-object interaction remain recognizable.\\
\textbf{Scene preservation} &
Whether the surrounding scene structure is preserved without unnecessary global changes.
\end{tabularx}

\vspace{4pt}
Use the following scale for each sub-score: 1 = failure, 2 = largely incorrect, 3 = partially correct, 4 = mostly correct, and 5 = clearly successful.

\vspace{4pt}
\textbf{Output} a concise rationale and the following numeric fields:

\vspace{2pt}
\begin{tabularx}{\linewidth}{@{}p{0.36\linewidth}X@{}}
\texttt{robotness} & integer score from 1 to 5\\
\texttt{target\_embodiment\_match} & integer score from 1 to 5\\
\texttt{interaction\_preservation} & integer score from 1 to 5\\
\texttt{scene\_preservation} & integer score from 1 to 5\\
\texttt{reasoning} & brief explanation
\end{tabularx}
\end{minipage}}
\end{table}

\subsection{Perceptual-Quality Prompt Template}

\begin{table}[H]
\centering
\caption{Perceptual-Quality prompt template used for VLM-based judgment. }
\label{tab:appendix_pq_prompt}
\setlength{\fboxsep}{7pt}
\fcolorbox{gray!35}{gray!20}{%
\begin{minipage}{1\textwidth}
\textbf{Perceptual-Quality Prompt Template}

\vspace{2pt}
You are evaluating the perceptual quality of an edited image.

\vspace{4pt}
\textbf{You are given:}

\vspace{2pt}
\begin{tabularx}{\linewidth}{@{}p{0.04\linewidth}X@{}}
(1) & an edited image.
\end{tabularx}

\vspace{4pt}
Judge only the visual quality of the edited image. Do not evaluate whether the requested robot embodiment is correct, and do not evaluate whether the editing instruction has been fully satisfied. Focus on whether the image is visually coherent as a single image.

\vspace{4pt}
Please assign three 1--5 sub-scores:

\vspace{2pt}
\begin{tabularx}{\linewidth}{@{}p{0.31\linewidth}X@{}}
\textbf{Naturalness} &
Whether the edited image looks visually plausible as a single image.\\
\textbf{Artifact absence} &
Whether the image avoids visible artifacts such as duplicated fingers, broken geometry, halos, blur, or distorted robot parts.\\
\textbf{Local coherence} &
Whether boundaries, texture, lighting, and occlusion near the edited region are consistent with the surrounding scene.
\end{tabularx}

\vspace{4pt}
Use the following scale for each sub-score: 1 = very poor, 2 = poor, 3 = acceptable, 4 = good, and 5 = high perceptual quality.

\vspace{4pt}
\textbf{Output} a concise rationale and the following numeric fields:

\vspace{2pt}
\begin{tabularx}{\linewidth}{@{}p{0.36\linewidth}X@{}}
\texttt{naturalness} & integer score from 1 to 5\\
\texttt{artifact\_absence} & integer score from 1 to 5\\
\texttt{local\_coherence} & integer score from 1 to 5\\
\texttt{reasoning} & brief explanation
\end{tabularx}
\end{minipage}}
\end{table}

\subsection{Embodiment Reference Bank}\label{app:embodiment_description_bank}

HandEdit uses a fixed embodiment reference bank for the two evaluation tracks: the standalone-hand track and the hand-arm track. Each target embodiment is represented by both visual URDF renders and a short textual description. The visual bank contains canonical rendered views of the target robot and is used for URDF-reference consistency in the identity metric. The textual bank is generated once from the canonical reference render using a multimodal vision-language model (Gemini 3.1 Pro) and remains frozen throughout the benchmark. These descriptions summarize visually discriminative cues, including palm geometry, finger structure, fingertip appearance, material, color, and exposed mechanical components.

\setlength{\LTcapwidth}{\textwidth}
\newcolumntype{M}[1]{>{\raggedright\arraybackslash}m{#1}}
\newcolumntype{G}[1]{>{\centering\arraybackslash}m{#1}}
\newcommand{\renderframe}[3]{\begingroup\setlength{\fboxsep}{0pt}\fcolorbox{gray!35}{white}{\includegraphics[width=#2,height=#3]{#1}}\endgroup}
\newcommand{\handrender}[1]{\begin{minipage}[c][1.82cm][c]{1.98cm}\centering\renderframe{#1}{1.62cm}{1.62cm}\end{minipage}}
\newcommand{\armrender}[1]{\begin{minipage}[c][1.72cm][c]{2.32cm}\centering\renderframe{#1}{2.18cm}{1.57cm}\end{minipage}}
\newcommand{\handbankrow}[3]{\parbox[c][1.88cm][c]{2.15cm}{\raggedright\textbf{#1}} & \handrender{#2} & \parbox[c][1.88cm][c]{8.52cm}{\raggedright #3} \\}
\newcommand{\armbankrow}[3]{\parbox[c][1.82cm][c]{2.72cm}{\raggedright\textbf{#1}} & \armrender{#2} & \parbox[c][1.82cm][c]{7.70cm}{\raggedright #3} \\}

\subsubsection{Standalone hand group}

\setlength\LTleft{0pt}
\setlength\LTright{0pt}
{\footnotesize
\setlength{\tabcolsep}{4.0pt}
\renewcommand{\arraystretch}{1.04}
\begin{longtable}{@{}M{2.22cm}G{2.02cm}M{8.62cm}@{}}
\caption{Frozen embodiment descriptions for the standalone-hand comparison group.}\label{tab:appendix_hand_desc}\\
\toprule
\textbf{Embodiment} & \textbf{Render} & \textbf{Frozen description} \\
\midrule
\endfirsthead
\caption[]{Frozen embodiment descriptions for the standalone-hand comparison group (continued).}\\
\toprule
\textbf{Embodiment} & \textbf{Render} & \textbf{Frozen description} \\
\midrule
\endhead
\midrule
\multicolumn{3}{r}{\footnotesize Continued on next page}\\
\endfoot
\bottomrule
\endlastfoot
\handbankrow{Ability}{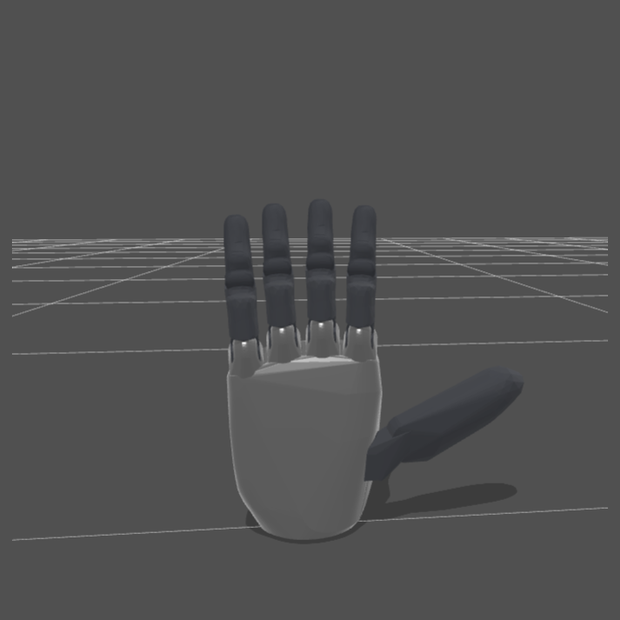}{A smooth white robotic hand with a rounded palm shell, dark rubber-like finger coverings, and a thick side-mounted thumb. The overall shape is compact and softly contoured, with minimal exposed mechanics and a clean cylindrical wrist base.}
\handbankrow{Allegro}{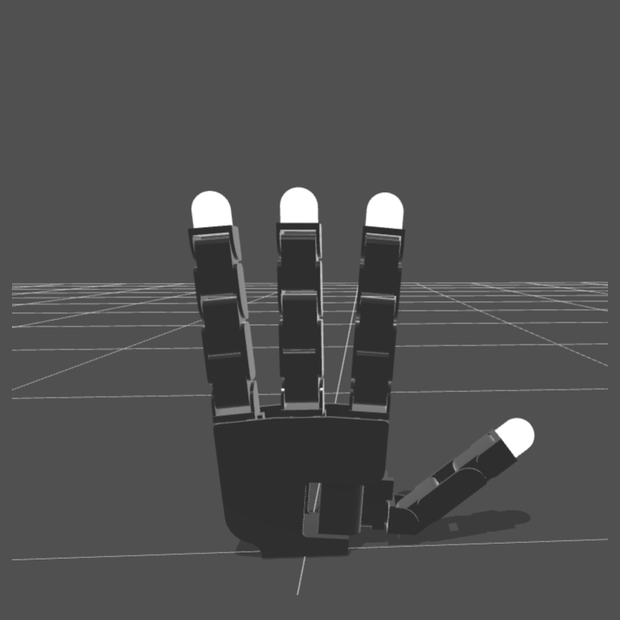}{A compact black robotic hand with a boxy palm, three straight front fingers made of rectangular joint modules, and a side thumb built from similar segmented black links. White rounded fingertip caps give the hand a high-contrast black-and-white appearance.}
\handbankrow{DexHand021}{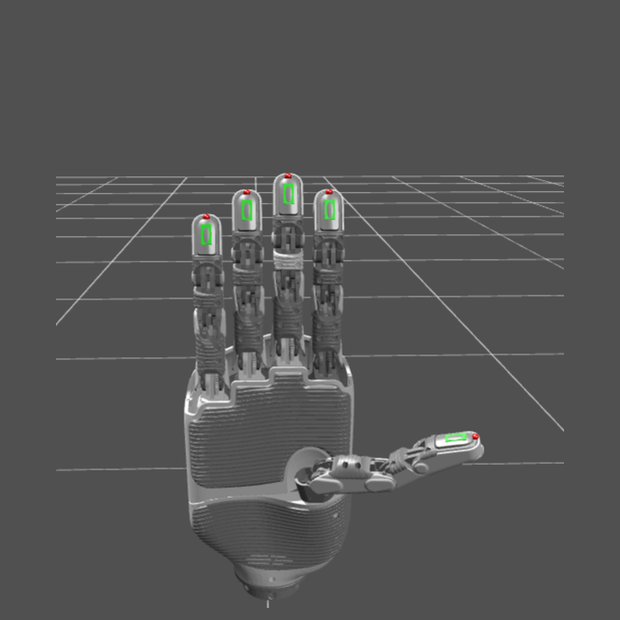}{A metallic dexterous hand with dense exposed mechanisms, a ribbed silver palm surface, and articulated fingers composed of visible links and cable-like structures. Bright green fingertip markers make the hand visually distinctive and strongly mechanical.}
\handbankrow{Leap}{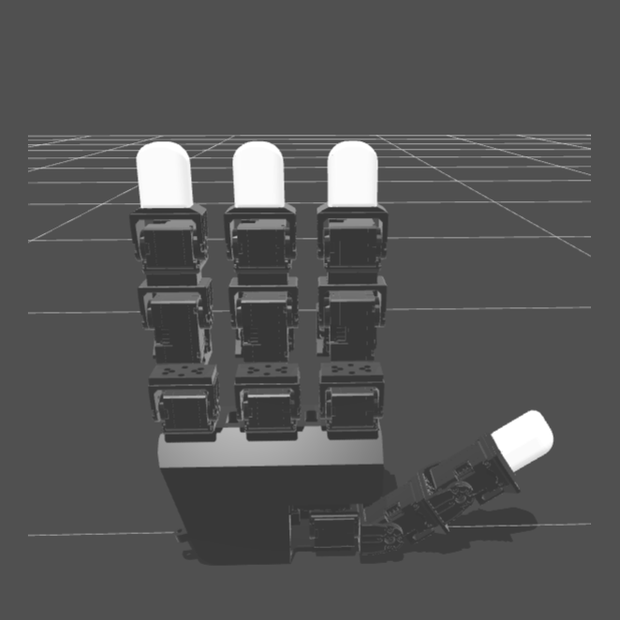}{A compact black robotic hand with a low-profile rectangular palm and fingers made of stacked box-like modules. The white fingertip caps and short angular side thumb give the hand a simple modular appearance.}
\handbankrow{OrcaHand}{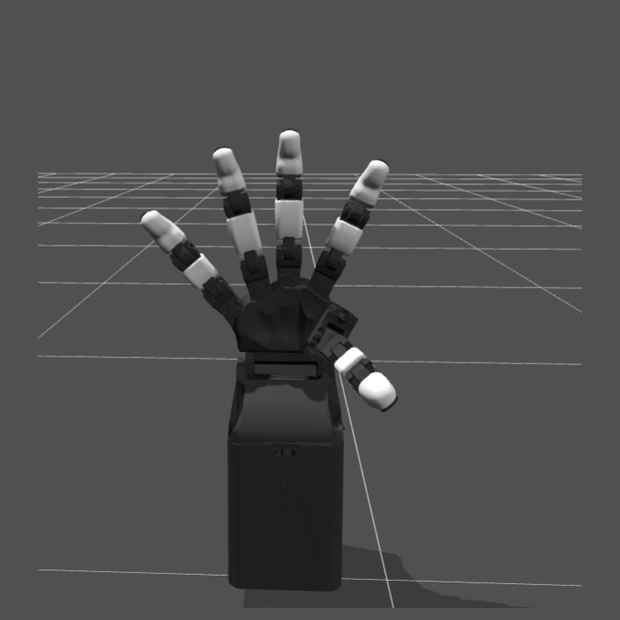}{A dark robotic hand with a narrow palm and five slender fingers spread in a fan-like configuration. Alternating dark segments and bright fingertip caps create an open, lightweight, highly splayed silhouette.}
\handbankrow{Revo2}{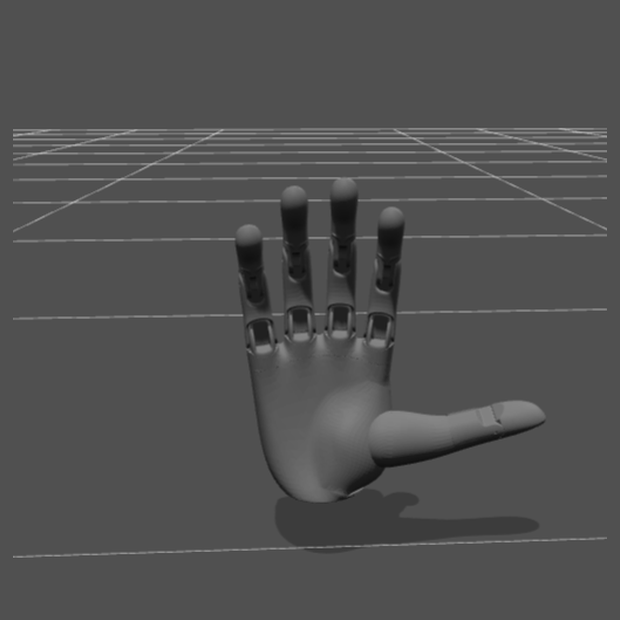}{A smooth anthropomorphic robotic hand with a rounded human-like palm, slim fingers, and softly curved matte-gray surfaces. The design has limited exposed mechanics and a naturalistic silhouette.}
\handbankrow{RH5DG2}{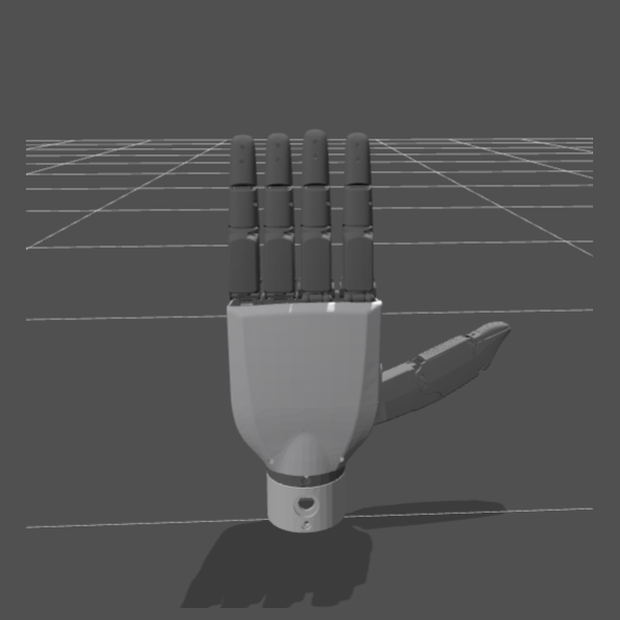}{A white-and-gray robotic hand with a rounded palm shell, dark cylindrical finger coverings, and a thick side thumb aligned close to the palm plane. It has an enclosed appearance with little exposed structure.}
\handbankrow{RH56DFX}{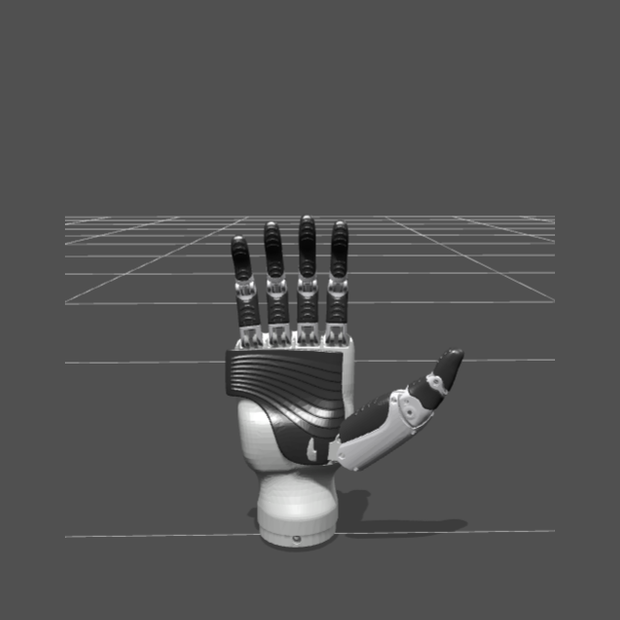}{A white-and-silver robotic hand with a sculpted palm shell, black finger pads, and a diagonal striped panel across the back of the palm. The large articulated side thumb and metallic details are visually distinctive.}
\handbankrow{RoHand}{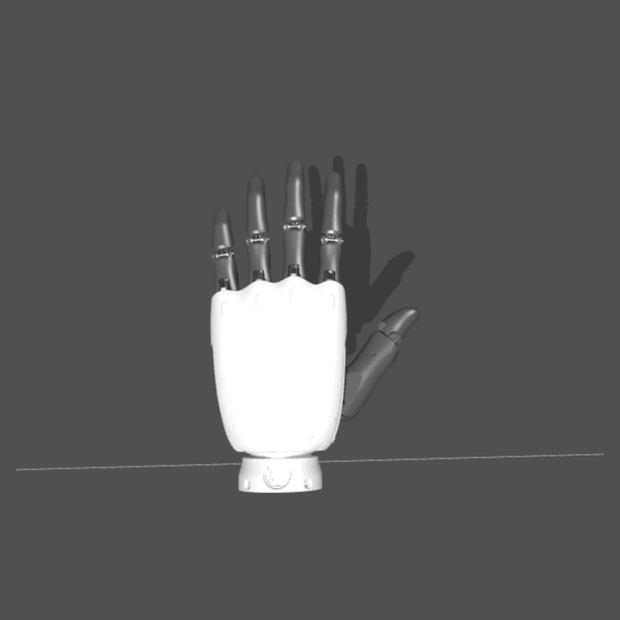}{A white robotic hand with a smooth enclosed palm shell, dark metallic finger segments, and a side-mounted thumb. Elongated anthropomorphic fingers and a compact rounded wrist base give it a polished integrated appearance.}
\handbankrow{Schunk SVH}{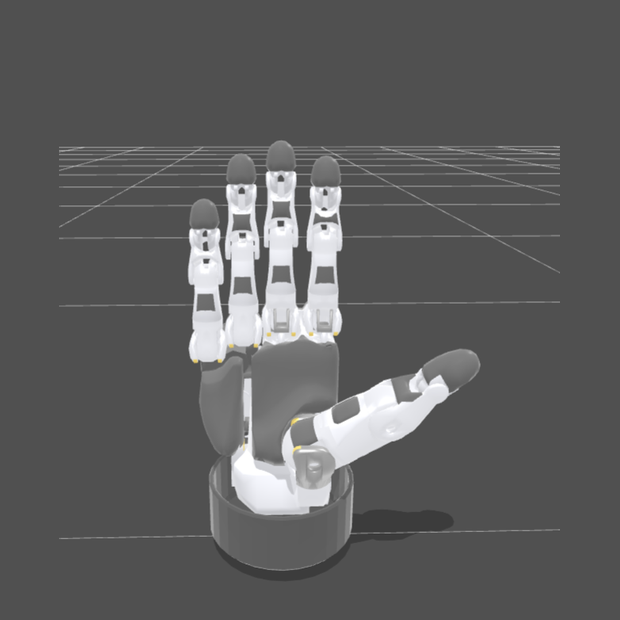}{A white dexterous hand with gray fingertip pads, robust finger links, and a dark central palm pad. The circular wrist base and thick articulated thumb give the hand a sturdy industrial appearance.}
\handbankrow{Shadow Hand}{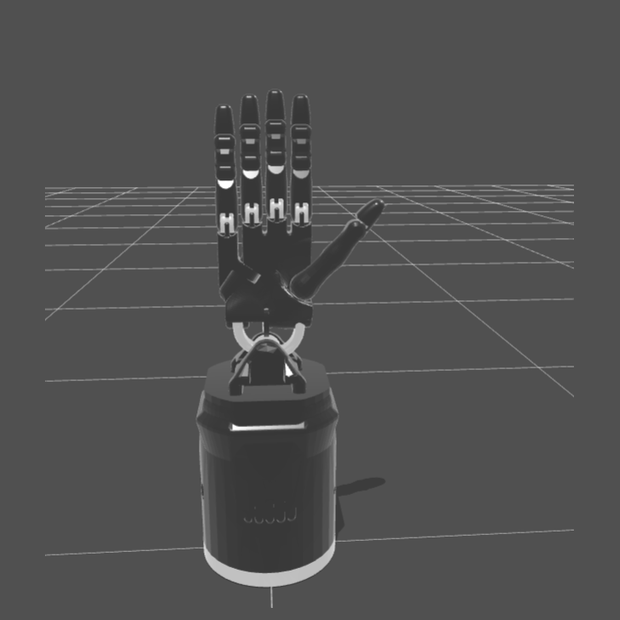}{A black anthropomorphic robotic hand with slim multi-joint fingers, white joint markings, and a curved white support structure near the wrist. The long human-like proportions make it dexterous but lightweight.}
\handbankrow{Sharpa}{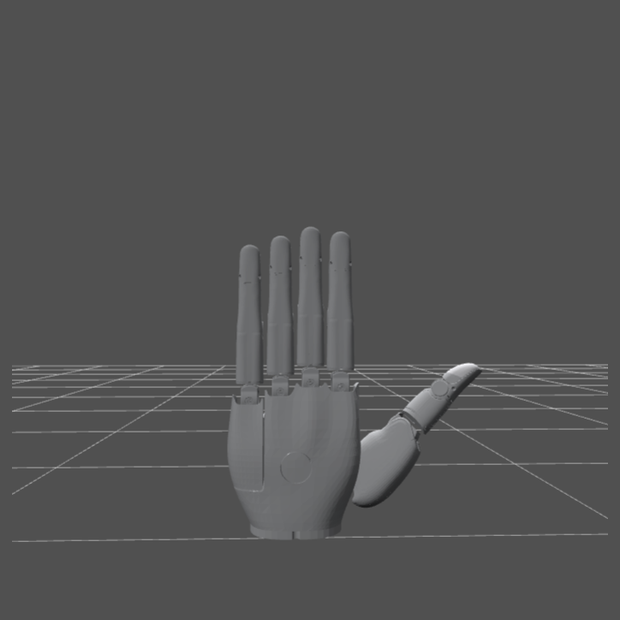}{A matte gray robotic hand with a clean anthropomorphic silhouette, smooth palm surfaces, and simple segmented fingers. The large articulated thumb contrasts with the otherwise minimal enclosed design.}
\handbankrow{Wuji}{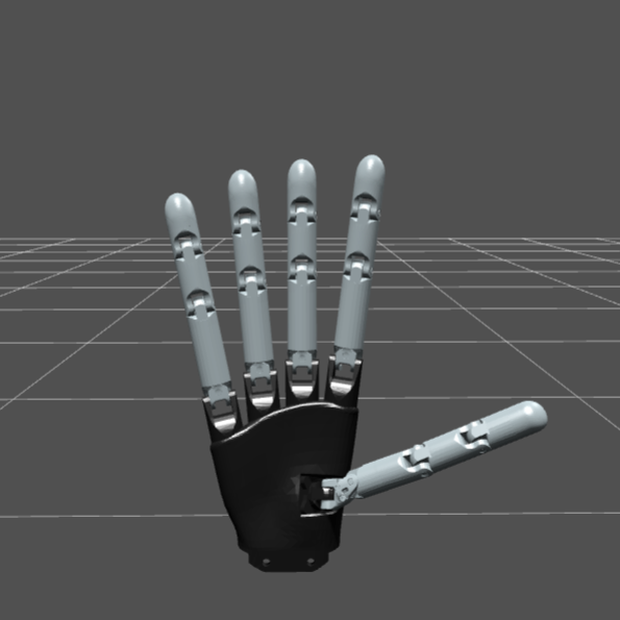}{A black-and-silver robotic hand with a glossy dark palm base, long light-gray articulated fingers made of segmented cylindrical links, and a widely extended side thumb. }
\end{longtable}}

\subsubsection{Hand-arm group}

{\footnotesize
\setlength{\tabcolsep}{4.0pt}
\renewcommand{\arraystretch}{1.04}
\begin{longtable}{@{}M{2.80cm}G{2.36cm}M{7.78cm}@{}}
\caption{Frozen embodiment descriptions for the hand-arm comparison group.}\label{tab:appendix_handarm_desc}\\
\toprule
\textbf{Embodiment} & \textbf{Render} & \textbf{Frozen description} \\
\midrule
\endfirsthead
\caption[]{Frozen embodiment descriptions for the hand-arm comparison group (continued).}\\
\toprule
\textbf{Embodiment} & \textbf{Render} & \textbf{Frozen description} \\
\midrule
\endhead
\midrule
\multicolumn{3}{r}{\footnotesize Continued on next page}\\
\endfoot
\bottomrule
\endlastfoot
\armbankrow{Jaka Zu7 + DexHand021}{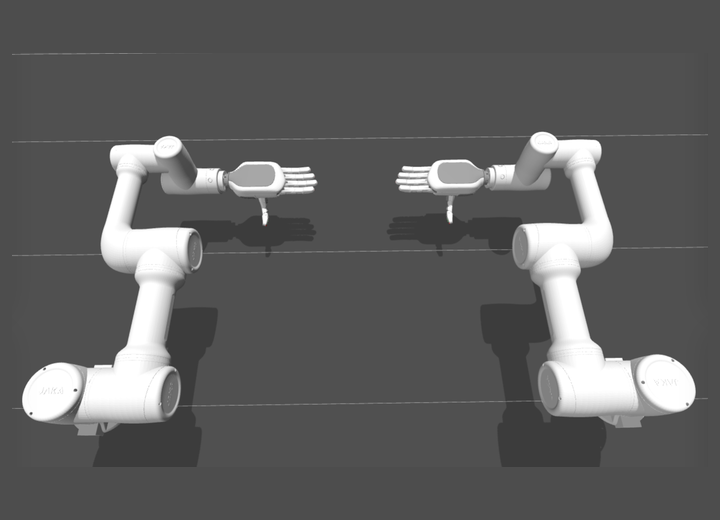}{A metallic-gray collaborative arm with blue circular joint caps, paired with a silver dexterous hand with dense exposed mechanisms, a ribbed palm surface, and slender articulated fingers. The transparent wrist housing and mechanical finger structure make it distinctive.}
\armbankrow{KUKA + Sharpa}{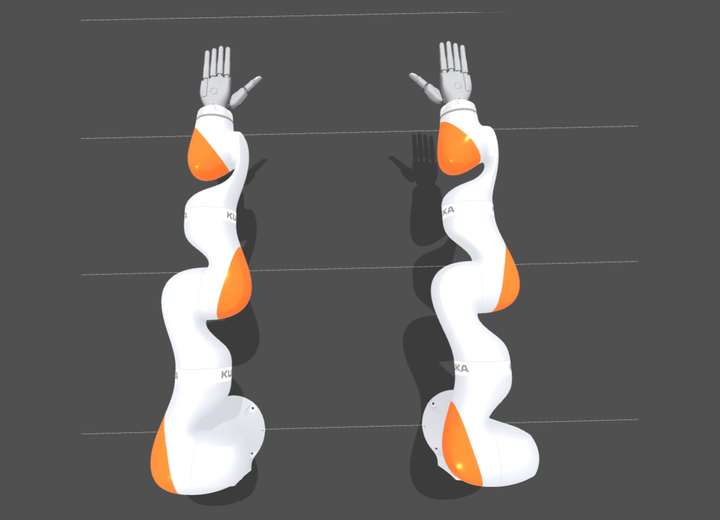}{A white KUKA-style arm with bright orange accent panels, paired with a smooth anthropomorphic hand. The clean enclosed hand silhouette contrasts with the strong white-orange arm styling.}
\armbankrow{Panda + Allegro}{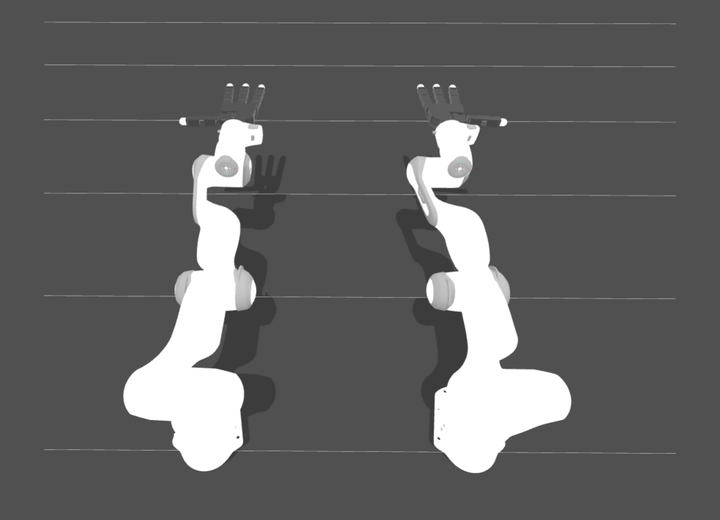}{A light Panda-style arm paired with a compact dark modular hand. The boxy palm, rectangular segmented fingers, and bright fingertip caps create a sharp contrast with the soft industrial arm.}
\armbankrow{Panda + Orca}{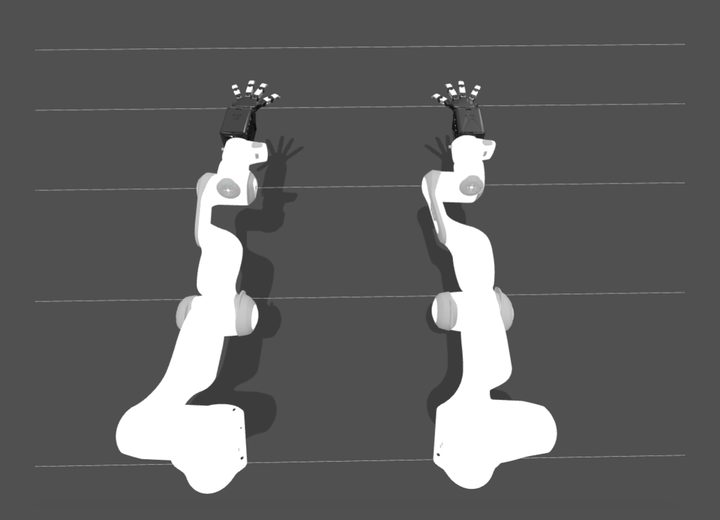}{A light Panda-style arm attached to a dark hand with a narrow palm and strongly spread fingers. Long separated digits with bright caps create an open fan-like silhouette.}
\armbankrow{RM65 + BrainCo}{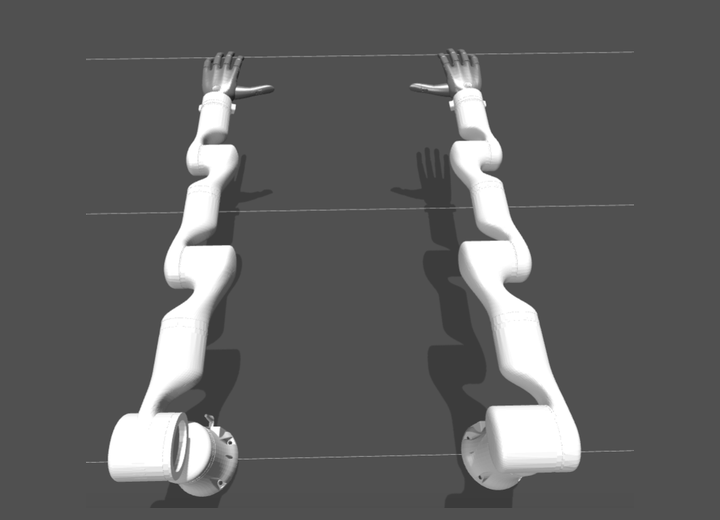}{A smooth white arm paired with a compact anthropomorphic hand with a rounded palm shell, slim fingers, and a clean enclosed structure. The embodiment appears polished and tightly integrated.}
\armbankrow{RM75 + RoHand}{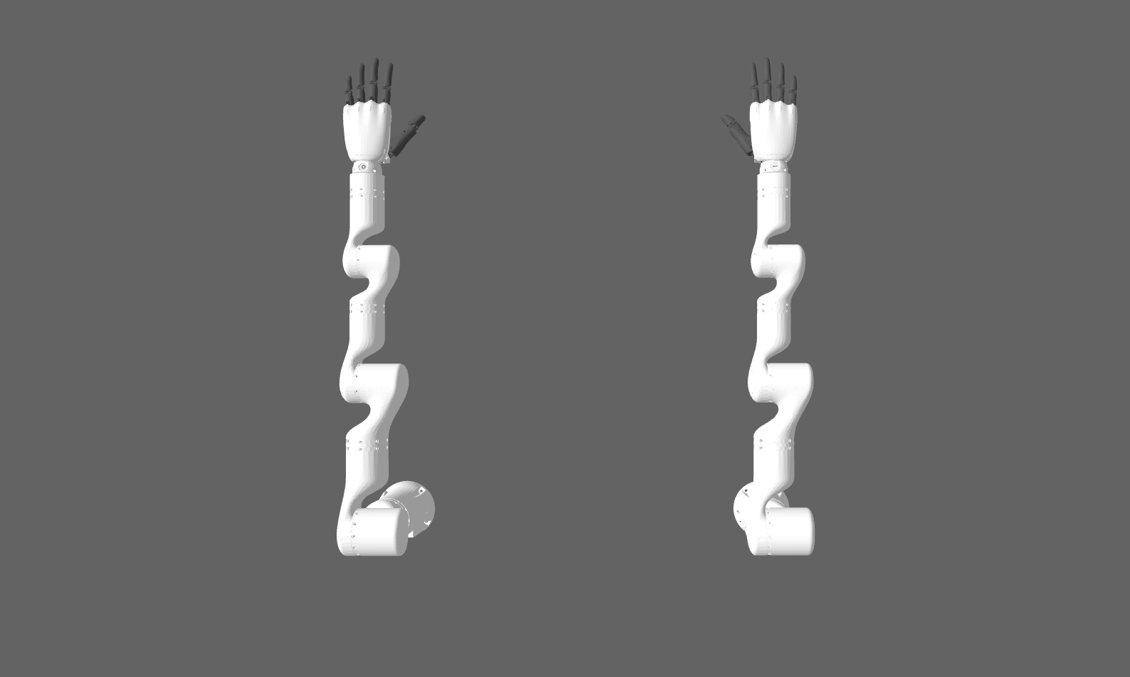}{A smooth white RM75-style arm with curved segmented links and rounded joint housings, paired with a white enclosed RoHand with dark finger segments and a side-mounted thumb. }
\armbankrow{UR5 + RH56DFX}{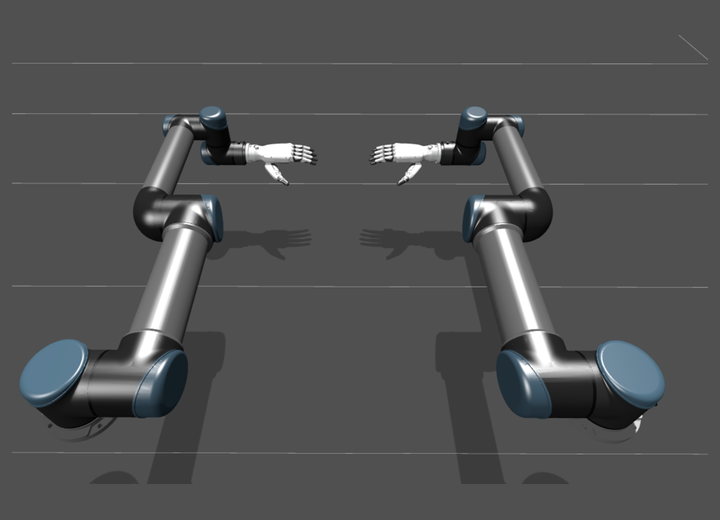}{A metallic-gray industrial arm with rounded cylindrical links and blue joint caps, ending in a white-and-silver dexterous hand with a sculpted palm shell, dark finger pads, and a large side thumb.}
\armbankrow{UR5 + RH5DG2}{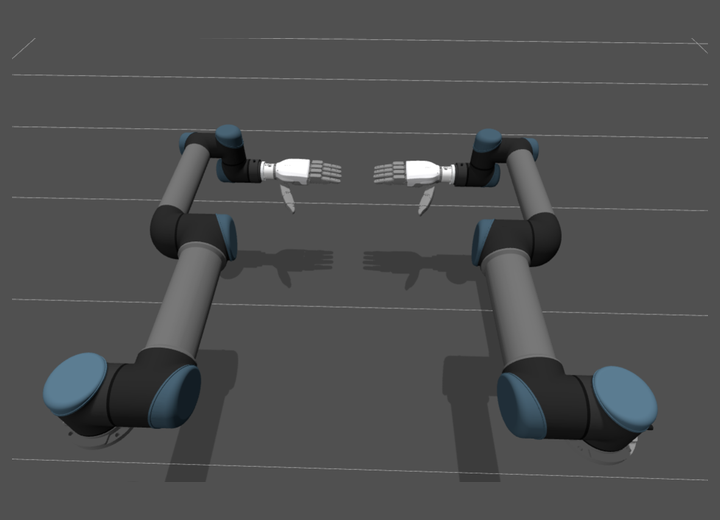}{A metallic-gray UR5-style arm with blue circular joint caps, attached to a white-and-gray hand with a rounded palm shell, dark cylindrical finger coverings, and a thick side thumb close to the palm plane.}
\armbankrow{UR5 + Schunk Hand}{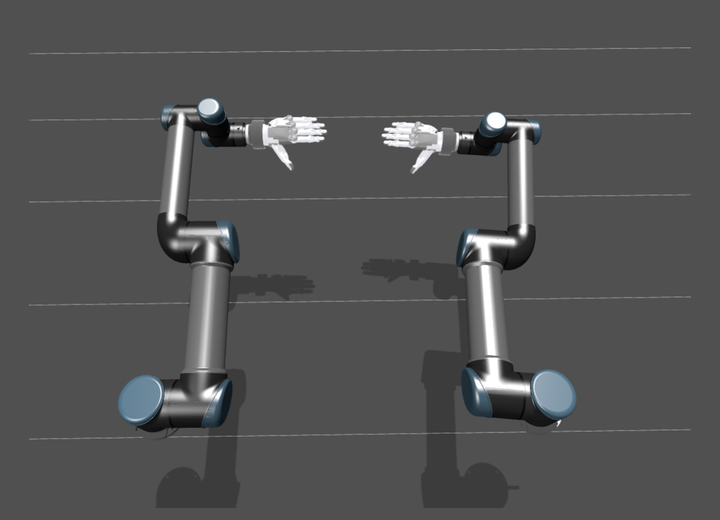}{A metallic-gray arm with blue round joint covers, paired with a robust white industrial hand. A broad palm, thick segmented fingers, gray fingertip pads, and heavy thumb create a sturdy precision-oriented look.}
\armbankrow{UR5 + Shadow Hand}{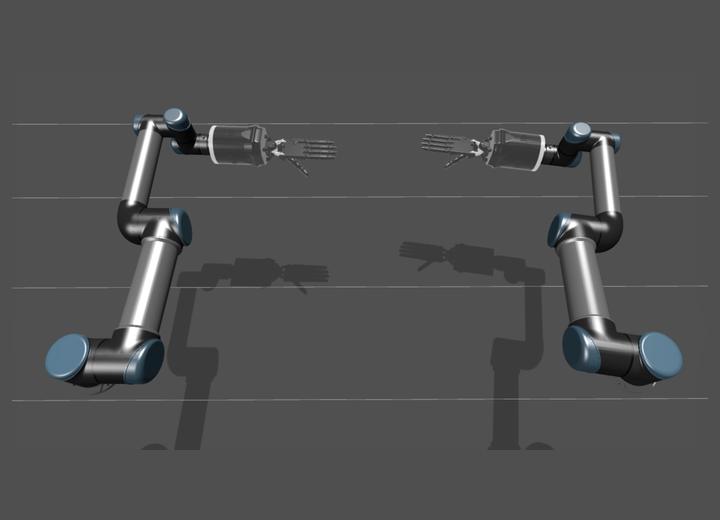}{A metallic-gray arm with blue circular joints, attached to a dark anthropomorphic hand with slim multi-joint fingers and bright fingertip caps. The hand is lightweight and human-like relative to the heavier arm.}
\armbankrow{UR5 + Wuji}{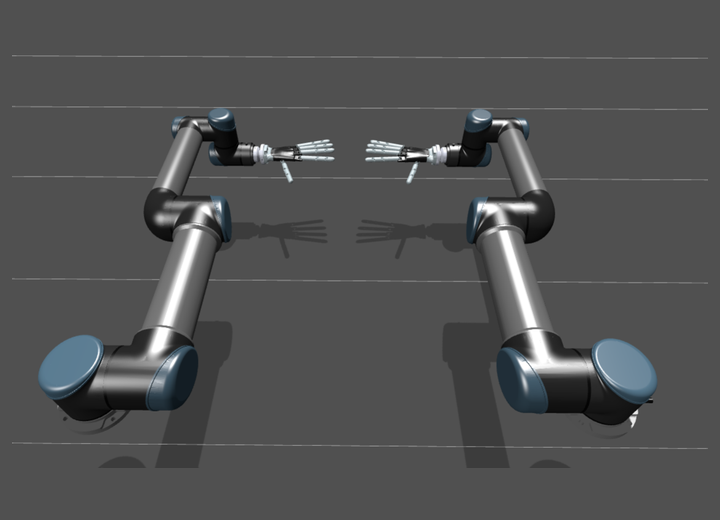}{A metallic-gray arm with blue circular joint caps, ending in a slim dexterous hand with long narrow fingers and a thin side thumb. The hand appears lighter and more elongated than other UR5-based embodiments.}
\armbankrow{xArm + Ability}{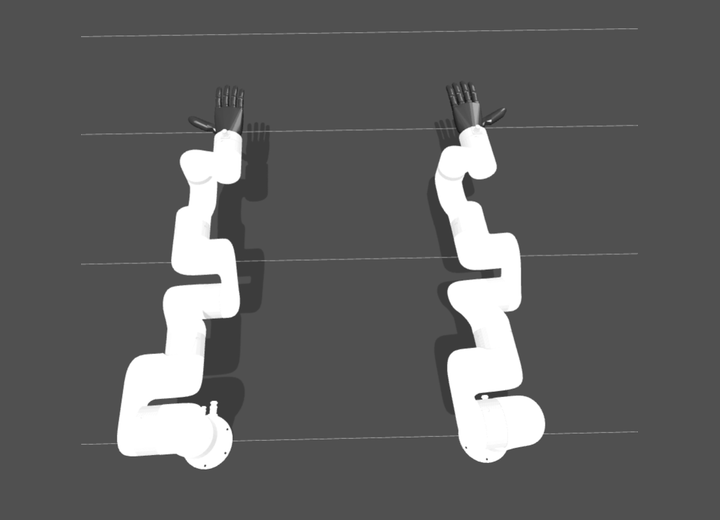}{A white arm with smooth enclosed links, paired with a compact white hand with a rounded palm shell, dark finger coverings, and a thick side thumb. The embodiment is soft-contoured and highly enclosed.}
\armbankrow{xArm + Leap}{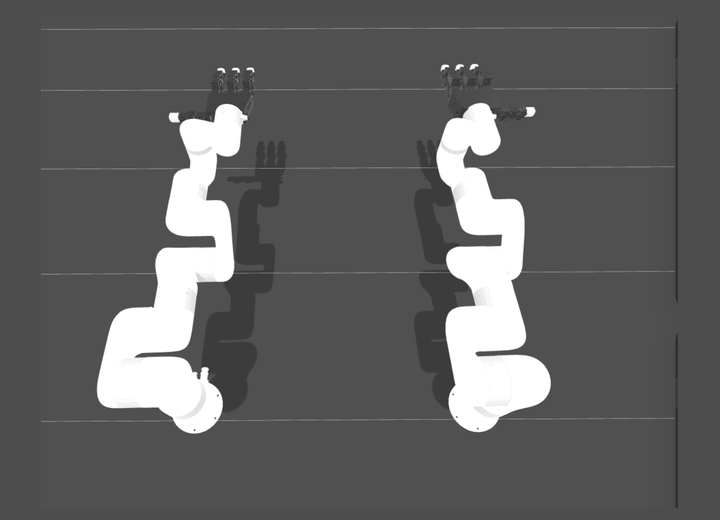}{A white arm with smooth rounded links, attached to a compact low-profile robotic hand with a blocky palm and modular fingers. Stacked box-like segments and bright caps give it a clean geometric appearance.}
\end{longtable}}

\section{Benchmark Model Details}\label{app:benchmark_model}
\begin{figure}[H]
\centering
\includegraphics[width=1\textwidth]{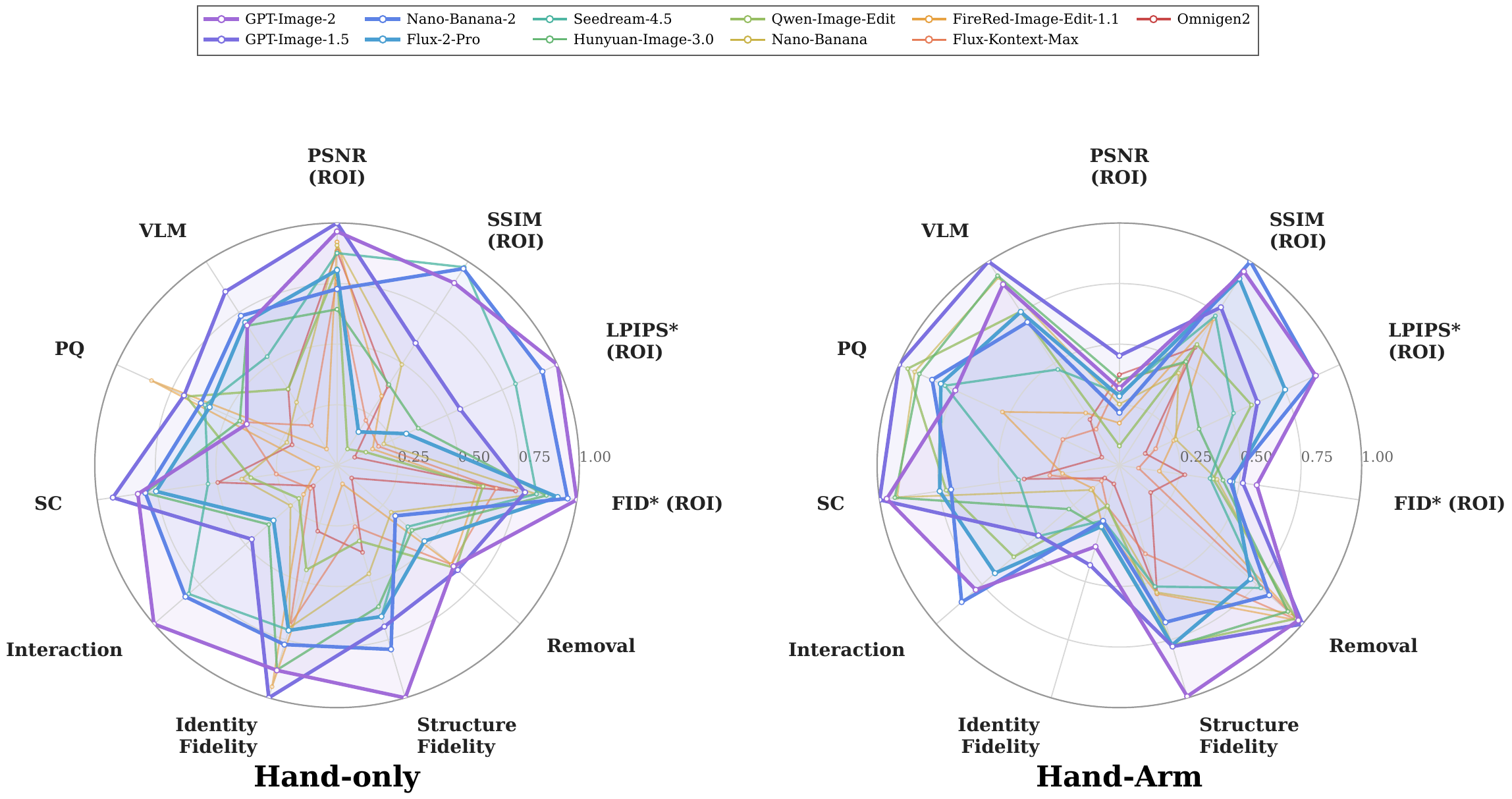}
\caption{\textbf{Radar chart comparison on the Hand-only and Hand-Arm tracks.} 
We visualize normalized scores across generic similarity metrics, embodiment-aware metrics, and VLM-based judgment. Metrics marked with $*$ are originally lower-is-better metrics; their scores are inverted for visualization so that larger values consistently indicate better performance.}
\label{fig:radar_results}
\end{figure}

As shown in Table~\ref{tab:baseline_roster} and Figure~\ref{fig:radar_results}, we evaluate a diverse set of commercial API-based and open-source image editing models. For each baseline, we report its model category, access mode, official input interface, and either the API cost for commercial models or the model size for locally deployed models.

\begin{table}[H]
\centering
\caption{\textbf{Evaluated baselines and their official input interfaces.} All models are evaluated using the default inference configuration.}
\label{tab:baseline_roster}
\resizebox{\textwidth}{!}{
\begin{tabular}{llllc}
\toprule
\textbf{Model} & \textbf{Family} & \textbf{Access} & \textbf{Input} & \textbf{Price/Model Size} \\
\midrule
\multicolumn{5}{@{}l}{\textbf{\textit{Commercial API-based Editors}}} \\
GPT-Image-2~\cite{openaiIntroducingChatGPT} & General Generative Model & API & Images + Text & \$211.0 /1k Imgs \\
GPT-Image-1.5~\cite{openaiChatGPTImages} & General Generative Model & API & Images + Text & \$133.0 /1k Imgs \\
Nano-Banana~\cite{nanobananaNanoBanana} & General Generative Model & API & Images + Text & \$39.0 /1k Imgs \\
Nano-Banana 2~\cite{geminiNanoBanana} & General Generative Model & API & Images + Text & \$67.0 /1k Imgs \\
FLUX-Kontext-Max~\cite{labs2025flux} & Editing Model & API & Images + Text & \$80.0 /1k Imgs \\
FLUX-2-Pro~\cite{flux-2-2025} & General Generative Model & API & Images + Text & \$45.0 /1k Imgs \\
Seedream-4.5~\cite{seedream2025seedream} & Editing Model & API & Images + Text & \$40.0 /1k Imgs \\
\midrule
\multicolumn{5}{@{}l}{\textbf{\textit{Open-source Editors}}} \\
Qwen-Image-Edit-2511~\cite{wu2025qwenimage} & Editing Model & Local & Images + Text & 20B Dense \\
FireRed-Image-Edit-1.1~\cite{firered2026rededit} & Editing Model & Local & Images + Text & 20B Dense \\
OmniGen2~\cite{wu2025omnigen2} & General Generative Model & Local & Images + Text & 7B Dense \\
Hunyuan-Image-3.0~\cite{cao2025hunyuanimage} & Editing Model & Local & Images + Text & 13B Active / 80B MoE \\
\bottomrule
\end{tabular}
}
\end{table}

\section{Benchmark Execution Details}\label{app:benchmark_execution}
The appendix specifies the standardized instruction templates used for benchmark evaluation. In the URDF-reference-conditioned protocol, each model receives the source image, a text instruction, and a  URDF render of the target embodiment. The render is used as an identity cue, specifying the target robot's morphology and appearance without providing an interaction pose. We also include a short visual description from Appendix~\ref{app:embodiment_description_bank} so that the textual prompt remains explicit and consistent across target embodiments.

\begin{table}[H]
\centering
\small
\caption{URDF-reference-conditioned instruction template.}
\label{tab:appendix_ref_prompt}
\begin{tabularx}{0.98\textwidth}{@{}L{2.7cm}X@{}}
\toprule
\textbf{Field} & \textbf{Instruction content} \\
\midrule
Input & Source image, target-robot render, and text instruction. \\
Embodiment cue & Target render, robot name, and compact visual descriptor from Appendix~\ref{app:embodiment_description_bank}. \\
Template & Replace the human [hand / hand-arm] region with the robot shown in the reference render: [robot name], described as [compact visual descriptor]. Preserve the original hand-object interaction, object state, and surrounding scene structure. \\
\bottomrule
\end{tabularx}
\end{table}

\section{Human Evaluation Details}
\label{app:human_eval}
\subsection{Annotation Setup}

To evaluate whether the proposed embodiment-aware metrics agree with human judgment on robot-hand editing quality, we conduct a blinded human annotation study on the test set. The study follows the same URDF-reference-conditioned setting as the benchmark. For each item, annotators are shown the source image, the text instruction, the target URDF render, and one edited output. The target render specifies the requested robot embodiment, but does not provide the desired interaction pose. Model names, metric scores, and pseudo-GT targets are hidden from annotators, preventing the study from becoming a pixel-level comparison against our composited target.

The study covers both the hand-only and hand-arm tracks and includes all 11 evaluated editors. We sample test images from the five source datasets and obtain 2,860 edited outputs in total. Each output is independently rated by five annotators on a five-point scale along three axes: \textbf{target-robot correctness}, \textbf{scene-and-interaction preservation}, and \textbf{realism with physical plausibility}. This gives 14,300 image-level annotations and 42,900 scalar ratings. The rating guide and the instruction shown to annotators are given in Tables~\ref{tab:appendix_human_axes} and~\ref{tab:appendix_human_prompt}.

\begin{table}[H]
\centering
\small
\caption{Rating guide for human evaluation. All axes use a five-point scale; the table shows the anchors for scores 1, 3, and 5.}
\label{tab:appendix_human_axes}
\setlength{\tabcolsep}{4pt}
\renewcommand{\arraystretch}{1.12}
\begin{tabularx}{0.98\textwidth}{@{}L{3.1cm}XXX@{}}
\toprule
\textbf{Axis} & \textbf{Score 1} & \textbf{Score 3} & \textbf{Score 5} \\
\midrule
Target-robot correctness
& The requested robot is missing, mostly human-like, or clearly the wrong embodiment.
& Robot cues are visible, but key morphology, material, or color details are wrong.
& The requested robot is clearly recognizable, with correct hand or hand-arm appearance. \\

Scene-and-interaction preservation
& The object, contact, occlusion, or scene layout is heavily changed.
& The main interaction is partly preserved, but visible contact or object-state errors remain.
& The original object state and hand-object interaction are well preserved. \\

Realism with physical plausibility
& Severe artifacts, broken geometry, or implausible scale/contact are present.
& The image is understandable, but artifacts or physical inconsistencies remain.
& The edit looks coherent, physically plausible, and visually natural. \\
\bottomrule
\end{tabularx}
\end{table}

\begin{table}[H]
\centering
\footnotesize
\caption{Instruction shown to human annotators.}
\label{tab:appendix_human_prompt}
\setlength{\fboxsep}{7pt}
\fcolorbox{gray!35}{gray!4}{%
\begin{minipage}{0.91\textwidth}
\textbf{Instruction Shown to Annotators}

\vspace{3pt}
You are evaluating an edited image for an egocentric human-to-robot hand editing benchmark.

\vspace{4pt}
For each example, you will see a source image, a text instruction, a reference render of the target robot, and one edited image. The reference render shows which robot should appear in the edited image. It does not show the exact interaction pose.

\vspace{4pt}
Please rate the edited image on three separate axes:
\begin{itemize}
\setlength{\itemsep}{1pt}
\setlength{\topsep}{2pt}
\item whether the requested target robot is correctly shown;
\item whether the original object, scene, and hand-object interaction are preserved;
\item whether the final image looks realistic and physically plausible.
\end{itemize}

\vspace{2pt}
Judge the edited image directly. Do not compare it to a hidden ground-truth image. Focus on whether the requested edit is successful and whether the final image remains coherent.
\end{minipage}}
\end{table}

\subsection{Human Evaluation Results}

For each output and each axis, we average the five annotator ratings to obtain a mean opinion score (MOS). We report Krippendorff's $\alpha$ to measure agreement on ordinal ratings. We also report ICC$(2,k)$, which measures the reliability of the averaged MOS under a two-way random-effects agreement model; here $k=5$ raters.

To compare human judgments with automatic metrics, we form pairwise model choices within each track. For every pair of editors, human preference is determined by the higher MOS, while metric preference is determined by the corresponding automatic score. We use ID Fidelity for target-robot correctness, Interaction for scene-and-interaction preservation, and the VLM judge score for realism with physical plausibility. Pairwise comparisons from the hand-only and hand-arm tracks are pooled, and Fleiss' $\kappa$ is reported in Table~\ref{tab:appendix_human_agreement}.

\begin{table}[H]
\centering
\small
\caption{Human agreement and metric-human agreement. }
\label{tab:appendix_human_agreement}
\begin{tabular}{lccc}
\toprule
\textbf{Evaluation} & \textbf{Target} & \textbf{Scene+Int.} & \textbf{Realism} \\
\midrule
Krippendorff's $\alpha$ among annotators & 0.624 & 0.603 & 0.627 \\
ICC$(2,k)$ among annotators & 0.846 & 0.831 & 0.849 \\
Metric-human agreement $\kappa$ & 0.586 & 0.572 & 0.554 \\
\bottomrule
\end{tabular}
\end{table}

The annotator agreement is reliable across the three axes, with Krippendorff's $\alpha$ above 0.60 and ICC$(2,k)$ above 0.83. The metric-human agreement is also consistent, with Fleiss' $\kappa$ ranging from 0.554 to 0.586. These results support our metric design, showing that the proposed automatic metrics are consistent with human judgments on target embodiment fidelity, interaction preservation, and visual plausibility.

\section{Limitations and Broader Impacts}
\label{app:limitations}
\textbf{Limitations.} Exact pixel-aligned human--robot image pairs cannot be obtained through physical capture because a human hand and a target robot embodiment cannot occupy identical configurations. HandEdit therefore uses edited pseudo-references as low-cost supervision. Such data retain a gap from real-robot observations in fine-grained appearance and contact dynamics. HandEdit is intended to provide scalable data for pre-training and mid-training and does not replace real-robot demonstrations; its use can reduce the amount of real-robot data required in subsequent stages. HandEdit also formulates embodiment editing as a well-defined image-editing task and supports the development of end-to-end editing models that reduce reliance on hand-crafted data-processing pipelines.

\textbf{Broader Impacts:}
HandEdit aims to provide a standardized testbed for evaluating human-to-robot dexterous hand image editing. We expect it to benefit both the embodied AI and image editing communities by supporting research on robot-centric data generation, dexterous embodiment transfer, and controllable image editing. By explicitly evaluating embodiment fidelity and hand-object interaction preservation, HandEdit can help identify failure modes that are not captured by generic image quality metrics alone.

\section{Visualization of More Examples}
\begin{figure}[H]
\centering
\includegraphics[width=1\textwidth]{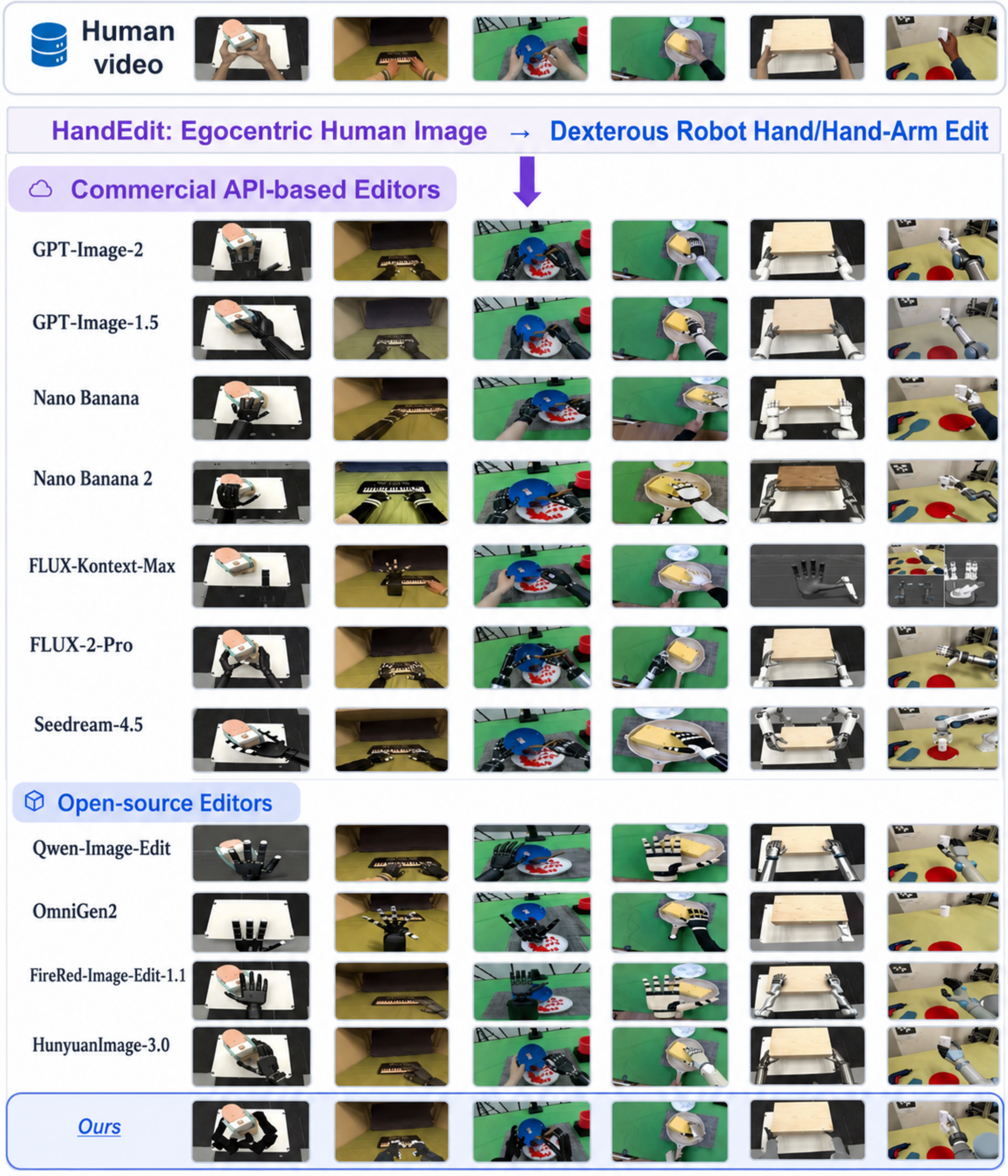}
\caption{\textbf{Comparison of Different Editing Methods on HandEdit.}}
\label{fig:results}
\end{figure}

See Figure~\ref{fig:results} for more data examples with various editing types in HandEdit.

\end{document}